\documentclass{article}
\usepackage{amssymb}

\usepackage[english]{babel}

\usepackage[letterpaper,top=2cm,bottom=2cm,left=3cm,right=3cm,marginparwidth=1.75cm]{geometry}
\usepackage{authblk}

\usepackage{amsmath}
\usepackage{graphicx}
\usepackage{xcolor}
\usepackage{algorithm}
\usepackage{algpseudocodex}
\usepackage[square, comma, sort&compress]{natbib}
\usepackage{lineno}
\usepackage{booktabs}
\usepackage{siunitx}
\usepackage[colorlinks=true, allcolors=black]{hyperref} % loaded last by convention

\date{}
\newcommand{\new}[1]{{\color{black}#1}}
\newcommand{\SIsec}[1]{SI, Section~#1}   % SI sections are numbered 1,2,3,4 (no S prefix)
\newcommand{\SIfig}[1]{SI, Fig.~S#1}
\newcommand{\SItab}[1]{Table~S#1}

\title{Eigenanalysis framework for autoregressive neural emulators of multi-scale chaotic dynamics}

\author{Conrad Ainslie$^{1}$, Pedram Hassanzadeh$^{2,3}$, Michael W. Mahoney$^{4,5,6}$ and Ashesh Chattopadhyay $^{1}$\thanks{aschatto@ucsc.edu}}
\affil{\normalsize $^1$Department of Applied Mathematics, University of California, Santa Cruz, Santa Cruz, 95064, CA}
\affil{\normalsize $^2$Department of Geophysical Sciences, University of Chicago, Chicago, IL 60637}

\affil{\normalsize $^3$Committee on Computational and Applied Mathematics, University of Chicago, Chicago, IL 60637}

\affil{\normalsize $^4$International Computer Science Institute, Berkeley, CA 94720}
\affil{\normalsize $^5$Lawrence Berkeley National Laboratory, Berkeley, CA 94720}
\affil{\normalsize $^6$Department of Statistics, University of California, Berkeley, CA 94720}

\begin{document}
\maketitle

\begin{abstract}
Neural autoregressive models have rapidly emerged as powerful emulators of high-dimensional chaotic systems, yet their long-term instability and error growth remain poorly understood, leading to ad-hoc solutions. Here, we develop an eigenanalysis framework that reveals the dynamical origin of this error growth. By analyzing the Jacobian of the learned one-step update map with respect to the state, we show how inference-time error growth, and thus model stability, is governed by its spectral radius. Direct-step architectures (models that predict the next state from the previous one) generically admit unstable eigenvalues with magnitudes exceeding one, explaining the rapid divergence of these widely used models. In contrast, integration-constrained models (where the time-derivative is estimated and integrated with a higher-order integrator) collapse their eigenspectrum onto the unit circle, yielding neutral stability and a universal linear error-scaling law. 
The largest eigenvalue of this Jacobian provides an architecture-agnostic, a priori diagnostic of short-term skill, long-term stability, and spectral bias, without requiring an expensive rollout. 
Leveraging this theory, we introduce a stability-promoting loss that explicitly regularizes Jacobian-driven error amplification, improving both forecast accuracy and dynamical robustness. Demonstrated across 29 models spanning two architectures, several explicit and implicit integrators, and multiple loss functions on the Kuramoto–Sivashinsky system, our results establish a theoretical foundation for the design and evaluation of neural emulators of chaotic multi-scale dynamics. More broadly, our framework is a step toward the kind of a priori stability analysis that numerical analysis provides for discretizations of differential equations, and that scientific machine learning currently lacks.
\end{abstract}
% Abstract word count: 248 (PNAS limit 250). Recount after any edit.

\section*{Significance Statement}
Neural networks that step a physical system forward in time now rival traditional simulations of chaotic, multi-scale flows such as the atmosphere, climate, and ocean, and they can run far faster than traditional numerical methods. 
They can also drift or blow up over long rollouts.
Remedies are currently found by trial and error, each tested by running the simulation that the network was built to replace. 
Here, we develop an eigenanalysis theory of these networks, which shows that one quantity, obtained from a single derivative of the trained model, predicts how fast its error grows, before any simulation is run. 
We find that networks that predict the next state directly amplify errors far faster than the physics allows, while networks that instead integrate a learned rate of change do not.

\section{Introduction}

Neural autoregressive models have become effective data-driven emulators of high-dimensional chaotic dynamical systems \citep{fan2020long,li2022learning,floryan2024instabilities}.
Their most visible success is in Earth system modeling, where they now rival or outperform operational numerical weather prediction \citep{pathak2022fourcastnet,lam2022graphcast,bi2022pangu,price2023gencast}.

Despite this progress, long-term emulation remains difficult.
Many models become unstable or excessively diffusive over long rollouts \citep{lippe2023pde,chattopadhyay2023challenges,keisler2022forecasting,lai2025machine,pedersen2025thermalizer}.

A growing number of models are now stable over long horizons, but this stability is typically achieved through extensive hyperparameter optimization; and even carefully tuned models can still blow up.
Each such failure tends to be addressed with a new, problem-specific fix rather than by identifying the mechanism that produced it, 
leaving the field without a principled account of when and why these models lose stability~\citep{watt2023ace,guan2024lucie,pedersen2025thermalizer,sambamurthy2025lazy}.

A leading explanation for this instability in multi-scale systems, such as turbulent flows with a decaying energy spectrum, is the \textit{spectral bias} of the network \citep{chattopadhyay2023challenges,yu2024tuning}, whereby a model trained to predict a single step fails to capture the high-wavenumber dynamics.
This error is thought to grow during autoregressive rollout through the coupling between small and large scales until it corrupts the resolved dynamics.
If spectral bias is the underlying cause, however, its consequences are realized only through the way one-step errors accumulate from step to step.
The role of this error propagation, as distinct from the one-step error itself, has not been made precise, and most stabilization strategies remain disconnected from any quantitative theory of how errors grow during rollout inference.

One line of work points to the integration scheme used to advance the model in time.
\citet{krishnapriyan2022learning} showed that constraining a model with a higher-order integrator, such as fourth-order Runge--Kutta, tends yield convergent integration analogous to classical numerical schemes.
\citet{chattopadhyay2023challenges} combined such a hard constraint with a spectral regularizer and reported reduced error growth and longer stable rollouts, and subsequent work has scaled this idea to coupled climate and high-resolution ocean models, producing stable rollouts over centuries of integration \citep{chattopadhyay2023oceannet,guan2024lucie,lupin2025simultaneous}.
Related stabilization strategies include refinement \citep{lippe2023pde}, noise-based regularization \citep{stachenfeld2021learned,wikner2022stabilizing}, and stochastic modeling \citep{chattopadhyay2022long,pedersen2025thermalizer,sambamurthy2025lazy}.

Despite this empirical evidence, it remains unclear why an integration constraint should help.
By analogy with numerical analysis, it is natural to assume that incorporating physical insight into a constraint inspired by time integration may aid stability; however, intuitions about combining physical reasoning and machine learning can be misleading \citep{failure21_TR,krishnapriyan2022learning,FalsePromizeZeroShot_TR}, and learned spatiotemporal models can behave counterintuitively \citep{annan2_understanding_bisases_TR}, including how performance scales with choice of the discrete time step \citep{bi2022pangu,chattopadhyay2021towards}.
A general theory of inference-time stability for these models is still lacking, with rigorous results largely confined to the linear case \citep{floryan2024instabilities}, and the effect of the integration scheme, loss function, and architecture on accuracy and stability has not been studied systematically.

This raises a set of questions.
\textcolor{black}{Does an integration constraint control the rate at which errors grow during rollout, and does it also improve short-term accuracy?}
Can a single quantity, computed before any long rollout, predict both?
Useful answers should not depend on the particular architecture, integration scheme, loss function, or system, and they should not require long emulations from many initial conditions.

\begin{figure}
\centering
\includegraphics[width=1.0\linewidth]{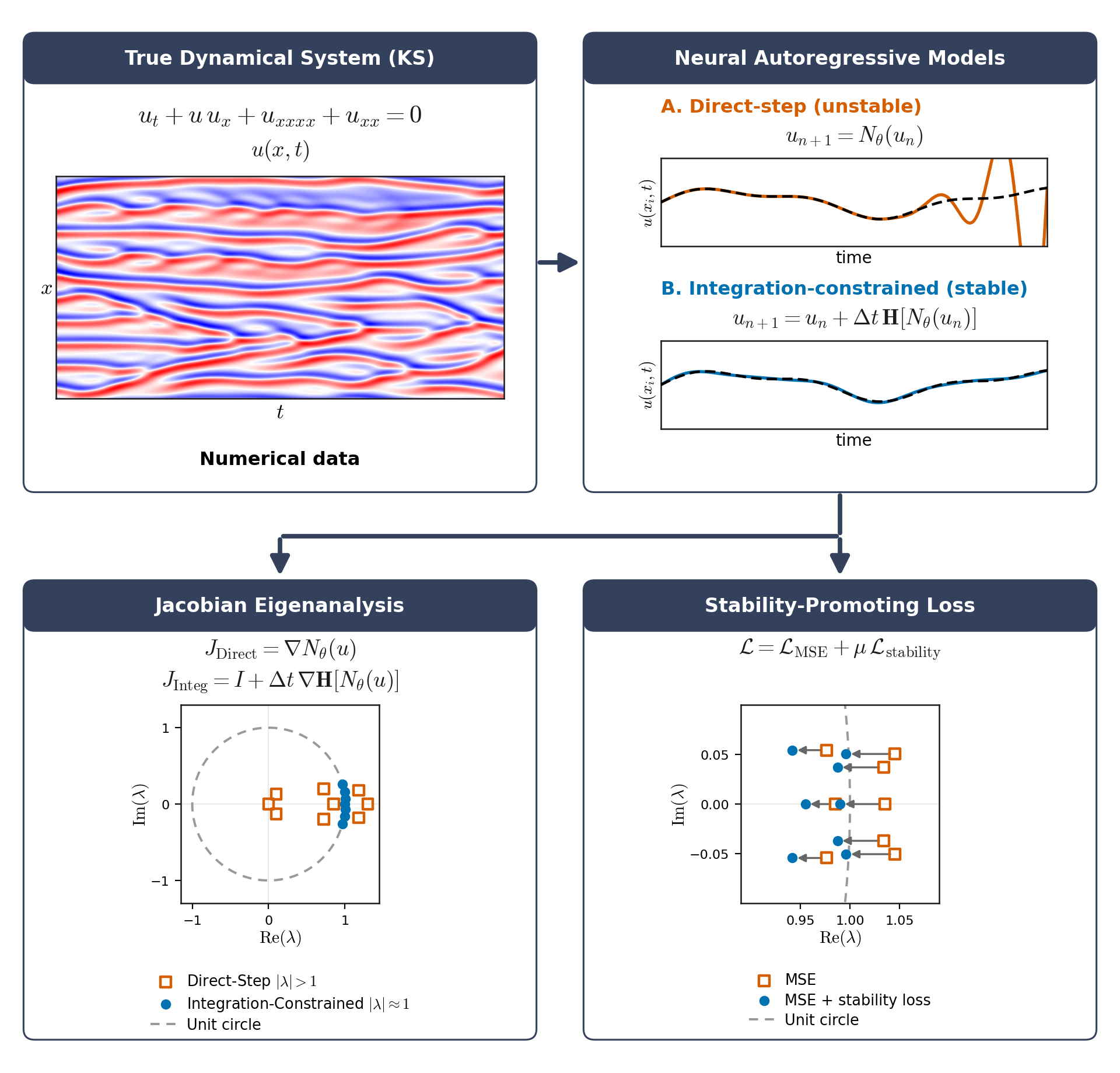}
\caption{\label{fig:Fig_7} \textbf{Schematic representation of the main contributions of this paper.} An eigenvalue decomposition of the Jacobian of the learned one-step update map, taken with respect to the system state and not with respect to the network parameters, yields a linear theory for the role the Jacobian plays in short-term error growth, \textcolor{black}{the rate of error accumulation over rollout,} and the spectral fidelity of the emulation. Direct-step models, which predict the next state in one shot, admit eigenvalues outside the unit circle and diverge during rollout, whereas integration-constrained models, in which the network estimates a time derivative that is advanced by a numerical integrator, collapse their eigenspectrum onto the unit circle and remain neutrally stable, so that a perturbation is neither amplified nor damped from one step to the next (Section~\ref{sec:eig_def}). 
The resulting stability-promoting loss regularizes the Jacobian-driven amplification of error, reducing the rollout error-growth rate: for some models this appears directly as eigenvalues pulled toward the unit circle (illustrated), while more generally it suppresses the realized, direction-dependent amplification of the error without necessarily changing $|\lambda_{\max}|$ (Section~\ref{sec:new_loss_results}).
}
\end{figure}

We answer these questions through an eigenanalysis of the Jacobian of the learned one-step update map (Fig.~\ref{fig:Fig_7}).
This Jacobian is taken with respect to the system state, which is the derivative that sets the stability of a discrete map in dynamical systems theory \citep{floryan2024instabilities}, and not with respect to the network parameters, which is the derivative returned by backpropagation and the one whose spectrum is usually studied in machine learning \citep{liao2021hessian}.
% VERIFY: liao2021hessian is a Hessian-spectrum reference. Confirm it is the
% intended citation for the parameter-side derivative, or replace it with a
% parameter-gradient reference already present in sample.bib.
The two derivatives have different arguments and different roles, and only the state derivative enters the error dynamics at inference.
Linearizing the error dynamics shows that inference-time error growth is governed by the spectral radius of this Jacobian, so its largest eigenvalue is an \textit{a priori} diagnostic that predicts error growth during inference.
This can be accomplished without resorting to expensive long rollouts, and it is independent of architecture, integration scheme, and loss function.
\textcolor{black}{This single quantity diagnoses the amplification of error during rollout; together with the one-step error, it accounts for short-term accuracy and the growth of spectral bias (Section~\ref{sec:deviation}).}
The analysis also explains the empirical record.
Direct-step models, which predict the next state directly, generically admit eigenvalues with magnitude well above one and have no mechanism to control them, which accounts for their rapid divergence.
Integration-constrained models, in which the network estimates a time derivative that is advanced by a numerical integrator, have a Jacobian whose eigenvalues collapse onto the unit circle and yield neutral stability together with a linear error-scaling law.
Building on this, we introduce a stability-promoting loss that directly regularizes the Jacobian-driven amplification of errors and improves both accuracy and stability.

We develop this theory for the Kuramoto--Sivashinsky (KS) system, and we evaluate it across the full suite of models studied here, spanning two architectures, several explicit and implicit integration schemes, and loss functions with and without spectral regularization.
The remainder of the paper presents the linear stability theory, the empirical results, the limits of the eigenvalue-only view and its distinctions from classical linear stability analysis, and the methods,  including all the explicit integration schemes and an implicit integration scheme based on implicit-layer and deep-equilibrium models \citep{kawaguchi2021theory}.
\new{The full model suite is listed in \SIsec{1}. The results obtained with the implicit constraint, together with the derivation of its Jacobian, are reported in \SIsec{2}. The Fourier-space analysis of spectral bias is given in \SIsec{3}, and the normality analysis that supports the eigenvalue approximation is given in \SIsec{4}.}

\section{Linear stability analysis}
\label{sec:theory}

We analyze the inference-time stability of a trained neural autoregressive model by linearizing its error dynamics during emulation, following the linear stability analysis of discrete dynamical systems.

\subsection{Learned dynamics and the Jacobian}
\label{sec:jacobian_def}

Let the true system advance one step through an operator $\mathbf{G}$,
\begin{equation}
    \mathbf{u}_T(t+\Delta t) = \mathbf{G}\left[\mathbf{u}_T(t)\right],
    \label{eq:true_dyn_sysm}
\end{equation}
with $\mathbf{u}_T$ the true state, usually available only at the initial condition $t=t_0$. A neural emulator approximates $\mathbf{G}$ by a learned operator $\tilde{\mathbf{G}}$ and advances its own predicted state $\mathbf{u}_p$,
\begin{equation}
    \mathbf{u}_p(t+\Delta t) = \tilde{\mathbf{G}}\left[\mathbf{u}_p(t)\right].
    \label{eq:pred_dyn_sys}
\end{equation}
Both states are functions of time. From here on we suppress the time argument whenever it is not needed and write $\mathbf{u}_T$ and $\mathbf{u}_p$, restoring it only where the rollout step matters.

The learned operator is built from a neural network $\mathcal{N}(\,\cdot\,;\theta)$ with parameters $\theta$ (see Section~\ref{sec:AR_models}). A direct-step model sets $\tilde{\mathbf{G}}[\mathbf{u}_T] = \mathcal{N}(\mathbf{u}_T;\theta)$, predicting the next state in one shot. An integration-constrained model instead estimates a time derivative with the network and advances it with a numerical integrator $\mathbf{H}$,
\begin{equation}
    \tilde{\mathbf{G}}[\mathbf{u}_T] = \mathbf{u}_T + \Delta t\,\mathbf{H}\!\left[\mathcal{N}(\mathbf{u}_T;\theta)\right].
    \label{eq:update_map}
\end{equation}
\new{Equation~(\ref{eq:update_map}) is the explicit form of the constraint. The implicit form, in which the network is evaluated at the updated state, is given in Section~\ref{sec:implicit_method}.}

The central object of our analysis is the Jacobian of the learned update map with respect to the state,
\begin{equation}
    \mathbf{J} = \nabla\tilde{\mathbf{G}}[\mathbf{u}_T].
    \label{eq:jac_def}
\end{equation}
Throughout, $\nabla$ denotes the gradient with respect to the state $\mathbf{u}_T$, evaluated at the current true state, and not the gradient with respect to the network parameters $\theta$. 
In machine learning, the derivative of a trained model most often refers to the derivative with respect to $\theta$, which is what backpropagation returns and what the optimizer acts on. Equation~(\ref{eq:jac_def}) is the other derivative, taken at fixed $\theta$, and it is the one that appears in the error recursion below.
The Jacobian is the sensitivity of the one-step prediction to a perturbation of its input, which is the quantity that governs how errors propagate during rollout. It is distinct from the parameter gradients used in training, and it is rarely examined when these models are evaluated.

\subsection{Error dynamics}
\label{sec:error_dynamics}

Define the prediction error $\mathbf{e}(t) = \mathbf{u}_T(t) - \mathbf{u}_p(t)$. Subtracting Eq.~(\ref{eq:pred_dyn_sys}) from Eq.~(\ref{eq:true_dyn_sysm}) gives
\begin{equation}
    \mathbf{e}(t+\Delta t) = \mathbf{G}\left[\mathbf{u}_T\right] - \tilde{\mathbf{G}}\left[\mathbf{u}_p\right].
    \label{eq:difference_bet_nn_sys}
\end{equation}
Linearizing $\tilde{\mathbf{G}}$ about the true state yields
\begin{equation}
    \mathbf{e}(t+\Delta t) = \underbrace{\mathbf{G}\left[\mathbf{u}_T\right] - \tilde{\mathbf{G}}\left[\mathbf{u}_T\right]}_{\boldsymbol{\epsilon}(t)} + \underbrace{\nabla\tilde{\mathbf{G}}\left[\mathbf{u}_T\right]}_{\mathbf{J}}\mathbf{e}(t) + \mathcal{O}\!\left(\|\mathbf{e}(t)\|_2^2\right).
    \label{eq:linearize}
\end{equation}
The first term $\boldsymbol{\epsilon}(t)$ is what we call the \textit{generalization error}, the one-step error of the learned operator at the true state, and the quantity that autoregressive models are trained to minimize. The second term, $\mathbf{J}\mathbf{e}(t)$, is what we call the \textit{propagated error}, the accumulated error carried forward through the linearized map, and $\mathbf{J}$ in this role is the error propagator. 
Training and validation control only $\boldsymbol{\epsilon}(t)$, whereas $\mathbf{J}$, which sets the size of the propagated error, is rarely examined.

At the initial condition the two error notions coincide. Since $\mathbf{u}_p(t_0)=\mathbf{u}_T(t_0)$, the first step incurs only the generalization error, so $\mathbf{e}(t_0+\Delta t)=\boldsymbol{\epsilon}(t_0)$. Writing $t_n = t_0 + n\Delta t$ and unrolling Eq.~(\ref{eq:linearize}) over $n$ steps, we obtain
\begin{align}
\mathbf{e}(t_n)
&= \boldsymbol{\epsilon}(t_{n-1})
+ \sum_{i=0}^{n-2}\left(
\prod_{j=i+1}^{n-1} \mathbf{J}(t_j)
\right)\boldsymbol{\epsilon}(t_i)
+ \mathcal{O}\!\left(\|\mathbf{e}(t_{n-1})\|_2^2\right),
\label{eq:autoreg_error_time}
\end{align}
with $\mathbf{J}(t_j)=\nabla\tilde{\mathbf{G}}[\mathbf{u}_T(t_j)]$ the Jacobian at step $j$. Total error growth therefore has two sources. The generalization error $\boldsymbol{\epsilon}$ reflects intrinsic model quality and can be reduced through better data, architecture, or capacity. 
The product of Jacobians is the error propagator over the intervening steps, and it determines whether those errors are amplified or damped, thus controlling stability during model rollout. 
Equation~(\ref{eq:autoreg_error_time}) cleanly separates model fidelity, carried by $\boldsymbol{\epsilon}$, from dynamical stability, carried by $\mathbf{J}$. Characterizing $\mathbf{J}$ across architectures, loss functions, and integration constraints, and relating it to model quality and stability during emulation, is the main contribution of this work.

\subsection{Eigenvalues and the Jacobian of each model class}
\label{sec:eig_def}

The Jacobian admits the eigendecomposition
\begin{equation}
    \mathbf{J} = \mathbf{V}\mathbf{\Sigma}\mathbf{V}^{-1},
    \label{eq:eigendec}
\end{equation}
with $\mathbf{V}$ the eigenvector matrix and $\mathbf{\Sigma}$ the diagonal matrix of eigenvalues $\lambda_1,\dots,\lambda_d$ for a system of dimension $d$. The magnitude of the largest eigenvalue $|\lambda_{\max}|$ sets the local stability. The dynamics are locally unstable when $|\lambda_{\max}|>1$, in which case a perturbation of the state is amplified at every step and grows geometrically in the step count, and neutrally stable when $|\lambda_{\max}|=1$, in which case a perturbation is neither amplified nor damped from one step to the next, so an error already present is carried forward at fixed size. For non-normal $\mathbf{J}$, as is typical of neural-network Jacobians, the eigenvalues bound the asymptotic growth of perturbations but not their transient amplification, which also depends on the orientation of the error relative to the eigenvectors; Section~\ref{sec:deviation} quantifies this effect. \textcolor{black}{\new{\SIsec{4}} shows that the integration constraint renders $\mathbf{J}$ near-normal, with a departure from normality of $\mathcal{O}(\Delta t^{2})$, so the eigenvalue approximation used below is valid.}

The two model classes have very different Jacobians. For a direct-step model,
\begin{equation}
    \mathbf{J} = \nabla\mathcal{N}(\mathbf{u}_T;\theta),
    \label{eq:jac_direct_step}
\end{equation}
which places no intrinsic constraint on the spectrum. 
On the other hand, differentiating the integration-constrained map of Eq.~(\ref{eq:update_map}) yields
\begin{equation}
   \mathbf{J} = \mathbf{I} + \Delta t\,\nabla\mathbf{H}\!\left[\mathcal{N}(\mathbf{u}_T;\theta)\right].
   \label{eq:jacobian_H}
\end{equation}
Because $\Delta t \ll 1$, the eigenvalues of Eq.~(\ref{eq:jacobian_H}) of an integration-constrained model stay close to one, and the model is near-neutrally stable by design. 
Direct-step models lack this control, since their Jacobian has no explicit dependence on $\Delta t$, and are thus prone to instability.
\new{Equation~(\ref{eq:jacobian_H}) is the Jacobian of the explicitly constrained map. Differentiating the implicit map instead gives $\mathbf{J} = (\mathbf{I} - \Delta t\,\nabla\mathbf{H}[\mathcal{N}])^{-1}$, which reduces to Eq.~(\ref{eq:jacobian_H}) to first order in $\Delta t$. \SIsec{2.2} derives this expression and shows that the inverse form places an eigenvalue outside the unit circle only when the learned tangent operator has an expanding direction.}

\new{The constraint fixes the form of $\mathbf{J}$ but not the size of the learned tangent operator $\nabla\mathbf{H}[\mathcal{N}(\mathbf{u}_T;\theta)]$, and the boundedness of that operator is a property the network acquires during training, rather than being an algebraic consequence of Eq.~(\ref{eq:jacobian_H}). For the true KS tangent operator at the resolution used here, $\Delta t\,k_{\max}^{4} = 1.07\times 10^{3}$, so an explicit Euler step applied to the exact linearization would be unstable by three orders of magnitude. The learned operators instead satisfy $\|\Delta t\,\nabla\mathbf{H}[\mathcal{N}]\| \sim 10^{-3}$, read from the spectral spread of Fig.~\ref{fig:Fig_1}(d). The near-unit spectrum therefore reflects the smoothness of the learned time derivative at high wavenumbers, which is the same property that produces the spectral bias analyzed in \SIsec{3}, and not the algebra of $\mathbf{I} + \Delta t\,\nabla\mathbf{H}$ alone.}

\subsection{Error scaling for integration-constrained models}
\label{sec:scaling_derivation}

We first derive the error growth over a single step, and we then extend it to $p$ steps for as long as the linear approximation holds.

From the linearized recursion of Eq.~(\ref{eq:linearize}), the error advances by one step as $\mathbf{e}(t_{k+1}) = \boldsymbol{\epsilon}(t_k) + \mathbf{J}(t_k)\mathbf{e}(t_k) + \mathcal{O}(\|\mathbf{e}(t_k)\|_2^2)$, with $\mathbf{J}(t_k)$ the Jacobian at step $k$. For an integration-constrained model $\mathbf{J}(t_k)=\mathbf{I} + \Delta t\,\nabla\mathbf{H}[\mathcal{N}(\mathbf{u}_T(t_k))]$ from Eq.~(\ref{eq:jacobian_H}), so
\begin{equation}
\mathbf{e}(t_{k+1}) = \boldsymbol{\epsilon}(t_k) + \left(\mathbf{I} + \Delta t\,\nabla\mathbf{H}\!\left[\mathcal{N}(\mathbf{u}_T(t_k))\right]\right)\mathbf{e}(t_k) + \mathcal{O}\!\left(\|\mathbf{e}(t_k)\|_2^2\right).
\label{eq:one_step_recursion}
\end{equation}

At the initial condition $\mathbf{u}_p(t_0)=\mathbf{u}_T(t_0)$, so $\mathbf{e}(t_0)=\mathbf{0}$ and the first step incurs only the generalization error, $\mathbf{e}(t_1)=\boldsymbol{\epsilon}(t_0)$. Applying Eq.~(\ref{eq:one_step_recursion}) once more, at $k=1$,
\begin{equation}
\mathbf{e}(t_2) = \boldsymbol{\epsilon}(t_1) + \left(\mathbf{I} + \Delta t\,\nabla\mathbf{H}\!\left[\mathcal{N}(\mathbf{u}_T(t_1))\right]\right)\mathbf{e}(t_1) + \mathcal{O}\!\left(\|\mathbf{e}(t_1)\|_2^2\right).
\label{eq:interim_scaling_expression}
\end{equation}
Because $\Delta t$ is small and $\tilde{\mathbf{G}}$ is continuous, the generalization error is nearly unchanged over one step, $\boldsymbol{\epsilon}(t_1)\approx\boldsymbol{\epsilon}(t_0)=\mathbf{e}(t_1)$. Substituting this into the first term gives
\begin{equation}
\mathbf{e}(t_2) = \left(2\mathbf{I} + \Delta t\,\nabla\mathbf{H}\!\left[\mathcal{N}(\mathbf{u}_T(t_1))\right]\right)\mathbf{e}(t_1) + \mathcal{O}\!\left(\|\mathbf{e}(t_1)\|_2^2\right).
\label{eq:final_scaling_expression}
\end{equation}
\new{The matrix acting on $\mathbf{e}(t_1)$ in Eq.~(\ref{eq:final_scaling_expression})} is $2\mathbf{I}$ plus a correction of order $\Delta t$, so for $\Delta t \ll 1$ its largest eigenvalue is close to $2$ and $\|\mathbf{e}(t_2)\| \approx 2\,|\lambda_{\max}|\,\|\mathbf{e}(t_1)\|$. This is due to the tight clustering of the spectrum. Since the eigenvalues of $\mathbf{J}$ are nearly coincident (Fig.~\ref{fig:Fig_1}(c,d)), $\|\mathbf{M}\mathbf{e}\| \approx |\lambda_{\max}(\mathbf{M})|\,\|\mathbf{e}\|$, where $\mathbf{M}$ is any square matrix with tightly clustered eigenvalues at unity, holds regardless of the orientation of $\mathbf{e}$, and the spectrum of $\left(2\mathbf{I} + \Delta t\,\nabla\mathbf{H}\!\left[\mathcal{N}(\mathbf{u}_T(t_1))\right]\right)$ coincides with $2\,|\lambda_{\max}|$ up to $\mathcal{O}(\Delta t)$. The error therefore doubles at the first comparison, which is the $k = 1$ reference line in Fig.~\ref{fig:Fig_2}(a). \new{\SIsec{4} shows that this directional insensitivity follows from the near-identity structure of Eq.~(\ref{eq:jacobian_H}) and not from the clustering of the eigenvalues alone.} Section~\ref{sec:deviation} examines the cases where this directional insensitivity breaks down. 

We now extend this to $p$ steps. Iterating Eq.~(\ref{eq:one_step_recursion}) from the initial condition and substituting the integration-constrained Jacobian gives
\begin{align}
\mathbf{e}(t_p)
&= \sum_{i=0}^{p-1}\boldsymbol{\epsilon}(t_i)
+ \Delta t\sum_{i=0}^{p-2}\sum_{j=i+1}^{p-1}
\nabla\mathbf{H}\!\left[\mathcal{N}(\mathbf{u}_T(t_{j}))\right]\boldsymbol{\epsilon}(t_i)
+ \mathcal{O}\!\left(\|\mathbf{e}(t_{p-1})\|_2^2\right) + \mathcal{O}\!\left(\Delta t^2\right),
\label{eq:autoreg_error_time_with_H}
\end{align}
whose leading term is the running sum of generalization errors. The same two approximations apply for as long as the rollout error stays small. The $\mathcal{O}(\Delta t^2)$ remainder and the explicit $\Delta t$ double sum are subdominant because $p\,\Delta t \ll 1$ over the step counts considered here, and the continuity of $\tilde{\mathbf{G}}$ keeps the generalization error nearly constant across the early rollout, $\boldsymbol{\epsilon}(t_i)\approx\boldsymbol{\epsilon}(t_0)=\mathbf{e}(t_1)$ for $i=0,\dots,p-1$. The leading sum then collapses to $p\,\mathbf{e}(t_1)$, and
\begin{equation}
\mathbf{e}(t_p) = \left(p\,\mathbf{I} + \Delta t\sum_{j=1}^{p-1} j\,\nabla\mathbf{H}\!\left[\mathcal{N}(\mathbf{u}_T(t_{j}))\right]\right)\mathbf{e}(t_1) + \mathcal{O}\!\left(\|\mathbf{e}(t_{p-1})\|_2^2\right) + \mathcal{O}\!\left(\Delta t^2\right).
\label{eq:p_step_scaling}
\end{equation}
The bracket is now $p\,\mathbf{I}$ plus a correction that is $\mathcal{O}(p\,\Delta t)$ relative to the leading term, so for $p\,\Delta t \ll 1$ its largest eigenvalue is close to $p$, giving the linear scaling law
\begin{equation}
\|\mathbf{e}(t_p)\| \approx p\,|\lambda_{\max}|\,\|\mathbf{e}(t_1)\|,
\label{eq:linear_scaling_law}
\end{equation}
with $|\lambda_{\max}|\approx 1$ the largest eigenvalue of the one-step Jacobian $\mathbf{J}$. The $2\mathbf{I}$ result of Eq.~(\ref{eq:final_scaling_expression}) is the $p=2$ case. The slope grows linearly with the step count $p$, reproducing the slopes $2$, $11$, $101$, and $1001$ measured at lead times $k = 1, 10, 100, 1000$ (that is, $p = 2, 11, 101, 1001$) in Fig.~\ref{fig:Fig_2}. Because $|\lambda_{\max}|\approx 1$, models separate along these lines only through the small departures of $|\lambda_{\max}|$ from one, so a larger $|\lambda_{\max}|$ corresponds to faster error growth.

The law holds only while both approximations remain valid, that is, while $\Delta t \ll 1$ and while $\|\mathbf{e}(t_p)\|$ stays small enough that the $\mathcal{O}(\|\mathbf{e}\|_2^2)$ terms are negligible. \textcolor{black}{Explicitly, the derivation requires $p\,\Delta t \ll 1$, i.e., $p \ll 1/\Delta t = 1000$ at $\Delta t = 10^{-3}$; the longer rollouts of Figs.~\ref{fig:Fig_1}(b) and \ref{fig:Fig_2}(d) probe beyond this window, and the continued slow accumulation observed there is an empirical observation outside the guaranteed regime rather than a prediction of Eq.~(\ref{eq:linear_scaling_law}).} \new{The two models trained at $\Delta t = 5\times 10^{-2}$ and $\Delta t = 10^{-1}$ exceed this window by a larger margin at the longest lead times, as discussed in \SIsec{2.4}.} As the rollout proceeds and the accumulated error grows, these higher-order terms become significant and the error departs from the linear prediction. \new{Within the lead times of Fig.~\ref{fig:Fig_2} this departure is visible first for the direct-step models, whose large per-step amplification excites the higher-order terms after only a few steps.}

\section{Results}
\label{sec:results}

We evaluate the Jacobian eigenanalysis of Section~\ref{sec:theory} as an \textit{a priori} diagnostic on the Kuramoto--Sivashinsky (KS) system. The suite studied here comprises 29 models, of which two are direct-step models and 27 are integration-constrained (see Methods). The main-text figures show a representative subset of these models, and the complete set of 29 is reported in \new{\SItab{1}}.

\subsection{\texorpdfstring{\textcolor{black}{Eigenvalues a priori diagnose the amplification of rollout error}}{Eigenvalues a priori diagnose the amplification of rollout error}}
\label{sec:linear_stability}

Figure~\ref{fig:Fig_1} contrasts the two model classes using four models, the direct-step MLP and FNO and their Euler-constrained counterparts. The two Euler models stand in for all 27 integration-constrained models examined here \new{(\SIfig{1} and \SIfig{2})}, and each model is trained and emulated at $\Delta t = \Delta t_{\text{DNS}}=10^{-3}$, unless stated otherwise. Throughout, a model whose Jacobian has the form of Eq.~(\ref{eq:jacobian_H}) is called integration-constrained, and every model except the two direct-step models is of this type. \new{For the three implicit models, this holds to first order in $\Delta t$ (\SIsec{2.2}).}

The central empirical result is in Fig.~\ref{fig:Fig_1}(a). The two direct-step models become unstable and unphysical within a few steps, with rapidly growing error, whereas the integration-constrained models remain stable and physically plausible, with error growing in proportion to the step count and orders of magnitude smaller. Figure~\ref{fig:Fig_1}(b) shows the same RMSE on log--log axes over \new{the first $10^4$ steps of the emulation}. The direct-step models saturate within a few tens to a few hundreds of steps, whereas the error of the two Euler models accumulates slowly and in proportion to the number of rollout steps, as predicted by Eq.~(\ref{eq:linear_scaling_law}), 
\new{remaining orders of magnitude smaller over most of the window, with the Euler MLP approaching the decorrelation level only at its right edge}. Strict proportionality to the step count does eventually give way as the accumulated error grows and the higher-order terms of Section~\ref{sec:scaling_derivation} become significant, but the error continues to accumulate slowly rather than diverging, in contrast to the direct-step models. \textcolor{black}{Equation~(\ref{eq:autoreg_error_time}) separates two budgets: the \new{eigenspectrum sets} the rate at which errors are amplified, while the one-step generalization error sets the rate at which they are injected.} The integration constraint alone, however, does not guarantee long-term stability. The near-neutral spectrum controls how fast errors are amplified, but the total error also carries the accumulated one-step generalization error $\boldsymbol{\epsilon}$ (Eq.~(\ref{eq:autoreg_error_time_with_H})). The Euler MLP of Fig.~\ref{fig:Fig_1}(b), whose $\boldsymbol{\epsilon}$ is the larger of the two, reaches the climatological decorrelation level by the right edge of that window, while the Euler FNO does not, and the same ordering holds for the remaining MLP and FNO pairs over the full $10^{5}$-step emulation \new{(\SIfig{1})}, despite eigenvalues on the unit circle in every case.
% VERIFY: confirm from the rollout data which MLP-based models reach the
% climatological decorrelation level within the 10^5-step emulation, and that
% no FNO-based model does. If any FNO model also saturates, weaken this to
% name the specific models rather than the architecture pairs.
Long-term stability therefore requires both a near-neutral spectrum and a sufficiently small generalization error, with or without the spectral regularizer (Section~\ref{sec:loss}), and the fidelity of the predicted spectrum still depends on architecture \new{(\SIsec{3})}.

This behavior is set by the eigenvalues of the Jacobian, evaluated at inference rather than during training, and computed once at the initial condition $t_0$. Figure~\ref{fig:Fig_1}(c) plots these eigenvalues on the Argand plane for all four models. The two direct-step models have \new{$|\lambda_{\max}| > 1$}, while the eigenvalues of the two Euler models collapse onto the unit circle and cluster tightly around $1$. The $2 \times 1024$ eigenvalues of the two integration-constrained models are nearly coincident there, so they appear as a single dense cluster and are barely visible in (c). Figure~\ref{fig:Fig_1}(d) therefore zooms into that cluster and shows \new{a spread confined to the third decimal place, with the two models separated in the fourth}, consistent with neutral stability.

\new{The measured values can be compared against the dynamics being emulated. The linear part of the KS operator has symbol $k^{2}-k^{4}$, whose maximum over the resolved wavenumbers is $0.2495$ at $k = 0.69$, so the corresponding time-$\Delta t$ flow map has $|\lambda_{\max}| = \exp(0.2495\,\Delta t) = 1.00025$. The departures from unity measured for the two Euler models, $1.75\times 10^{-4}$ and $2.01\times 10^{-4}$, agree with the value implied by the linear operator, $2.5\times 10^{-4}$, to within a factor of $1.5$. This estimate neglects the advective contribution to the tangent map and is a reference scale rather than a precise target. It matters because Eq.~(\ref{eq:jacobian_H}) would place the spectrum within $\mathcal{O}(\Delta t)$ of unity for any bounded learned operator, including one that has learned the wrong dynamics, so proximity to unity is not by itself evidence that the model has learned anything. Agreement to within a factor of $1.5$ with the value implied by the linear operator indicates instead that the trained network has recovered the size of the true tangent map and has not merely inherited the near-identity structure of the constraint. It also gives $|\lambda_{\max}|$ a reference value against which a trained model can be checked before any rollout is run, since a model whose measured $|\lambda_{\max}|$ departs from the value implied by the linear operator is misrepresenting the local growth rate of the system it emulates.}

This eigenvalue structure explains the error growth in (a) and (b). The near-unit eigenvalues of the integration-constrained models produce the slow, near-neutral accumulation derived in Section~\ref{sec:scaling_derivation}, while the large $|\lambda_{\max}|$ of the direct-step models (Eq.~(\ref{eq:jac_direct_step})) amplifies perturbations strongly and drives the rapid divergence. The collapse onto the unit circle follows directly from Eq.~(\ref{eq:jacobian_H}). The identity term contributes eigenvalues exactly at $1$, and the correction $\Delta t\,\nabla\mathbf{H}[\mathcal{N}(\mathbf{u}_T)]$ is a small perturbation because $\Delta t = 10^{-3}$. The qualitative spectrum, and hence the predicted neutral stability, is therefore insensitive to the architecture, the loss function, the presence or absence of spectral regularization, and the order of the explicit integrator. This insensitivity is why two Euler models can stand in for all 27 integration-constrained models in Fig.~\ref{fig:Fig_1}. \new{\SIfig{1} shows the RMSE curves and the eigenspectra of the remaining explicit models, which exhibit the same behavior.}

\subsection{Emergence of a linear scaling law}
\label{sec:linear_scaling}
Figure~\ref{fig:Fig_2} tests the scaling law of Section~\ref{sec:scaling_derivation} for all 29 models, plotting $\|\mathbf{e}(t+\Delta t)\|$ against $|\lambda_{\max}|\,\|\mathbf{e}(t)\|$ at successive rollout steps, with $|\lambda_{\max}|$ the largest eigenvalue of the Jacobian $\mathbf{J}$ evaluated at the initial condition. The two direct-step models are shown in vermilion and the 27 integration-constrained models---spanning the Euler, RK4, PEC4, and implicit Euler schemes---in blue, with architecture distinguished by marker fill (MLP filled, FNO open). \new{\SItab{1}} lists each model with its integration scheme, architecture, and loss.
% Figure~\ref{fig:Fig_2} tests the scaling law of Section~\ref{sec:scaling_derivation} by plotting $\|\mathbf{e}(t+\Delta t)\|$ against $|\lambda_{\max}|\,\|\mathbf{e}(t)\|$ at successive rollout steps, with $|\lambda_{\max}|$ the largest eigenvalue of the Jacobian $\mathbf{J}$ evaluated at the initial condition. The subset shown comprises the two direct-step models and the integration-constrained models built from the Euler, RK4, and PEC4 schemes with both the MLP and FNO architectures. The complete set of 26 models is reported in Appendix~\ref{app:all_models}, which also lists each model with its integration scheme, architecture, and loss.

The integration-constrained models follow the predicted law $\|\mathbf{e}(t_p)\| \approx p\,|\lambda_{\max}|\,\|\mathbf{e}(t_1)\|$ of Eq.~(\ref{eq:linear_scaling_law}). At rollout step $p$ they lie on a line whose slope is the step index $p$ itself, equal to 2 at the first step and growing to 11, 101, and 1001 at $p = 11, 101, 1001$; on the log--log axes of the four panels of Fig.~\ref{fig:Fig_2} these appear as unit-slope reference lines, one per panel, whose vertical offset grows with $p$. Because $|\lambda_{\max}|\approx 1$ for these models, they separate along each line only through the small departures of $|\lambda_{\max}|$ from one.

\new{The four panels of Fig.~\ref{fig:Fig_2} correspond to $p\,\Delta t = 0.002$, $0.011$, $0.101$, and $1.001$, so they traverse the validity window of Eq.~(\ref{eq:linear_scaling_law}) from well inside it to its boundary. The law holds tightly in (a) and (b), where $p\,\Delta t \ll 1$. The scatter about the reference line grows in (c), where $p\,\Delta t \approx 0.1$. In (d), where $p\,\Delta t \approx 1$, the condition is no longer met and the models depart from the line in both directions. The breakdown therefore occurs at the step count the derivation predicts, and the agreement in panels (a) to (c) is a test of the law within the regime where it is derived rather than an extrapolation beyond it.}

The two direct-step models track the same line for the first few steps and then peel away, growing faster than linearly. This is consistent with the law being a small-error prediction. The eigenvalues of the direct-step models are not clustered near one, so once the accumulated error grows enough to excite the dominant eigenvalue with \new{$|\lambda_{\max}| > 1$}, the higher-order terms take over. The departure occurs after about 10 steps for the direct-step MLP and about 100 steps for the direct-step FNO, \new{which places them above the reference line at the lead time of panel (b). Beyond that point, their error reaches the climatological decorrelation level and stops growing, so at the longer lead times of panels (c) and (d) they fall below a reference line that continues to grow in proportion to $p$. This saturation is the same one visible in Fig.~\ref{fig:Fig_1}(a).}

The slope $p$ has a simple origin (Eq.~(\ref{eq:p_step_scaling})). At the first step $\mathbf{e}(t_1)$ is the generalization error, and the Jacobians of integration-constrained models have $|\lambda_{\max}|\approx 1$ for $\Delta t \ll 1$, so the generalization error is nearly unchanged over the early rollout and Eq.~(\ref{eq:autoreg_error_time_with_H}) reduces to a sum of $p$ near-identical copies of $\mathbf{e}(t_1)$. \new{As $p$ grows, the accumulated error eventually becomes large enough that the $\mathcal{O}(\|\mathbf{e}\|_2^2)$ terms are no longer negligible, and the points scatter about the line, which is what panel (d) shows.}

Enforcing an integration constraint is what makes this linear stability analysis usable. The constraint forces the spectrum onto the unit circle and fixes a single controlled scale, $|\lambda_{\max}|\approx 1$, about which the error dynamics linearize cleanly. The direct-step models have no such control, their spectrum spans a wide range \new{with $|\lambda_{\max}| > 1$}, and no comparable linear law holds for them over the rollout.

\new{The same holds when the constraint is imposed with an implicit rather than an explicit integrator. Three of the models are constrained with an implicit Euler scheme, two of them trained and emulated at $\Delta t = 5\times 10^{-2}$ and $\Delta t = 10^{-1}$. At every $\Delta t$ tested, the implicit constraint gives a $|\lambda_{\max}|$ closer to unity and a lower rate of error growth than the corresponding explicit constraint, and the ordering of $|\lambda_{\max}|$ reproduces the ordering of the measured RMSE. This transfers the classical result that implicit integrators are more stable at large $\Delta t$ to the learned setting, and it follows from the inverse structure of the implicit Jacobian derived in \SIsec{2.2}. \SIsec{2} reports the full comparison.}

The largest eigenvalue is therefore an \textit{a priori} diagnostic of model behavior, obtained by differentiating the model output with respect to its input through automatic differentiation, without running a full rollout. A larger $|\lambda_{\max}|$ implies faster error growth, hence poorer short-term accuracy and weaker long-term stability. This ranking holds both across the two model classes and within the integration-constrained class, where the small differences in $|\lambda_{\max}|$ order the models by error growth along the lines of Fig.~\ref{fig:Fig_2}. Section~\ref{sec:deviation} quantifies this predictive power with a linearized error law, built from $|\lambda_{\max}|$ and the measured one-step error, that reproduces the measured rollout error across the explicit integration-constrained suite (Fig.~\ref{fig:Fig_4}(c,d)).
\new{Because $|\lambda_{\max}|$ is a per-step multiplier, models trained at different time steps are compared through the growth rate per unit time $\sigma = \ln|\lambda_{\max}|/\Delta t$. Measured this way, every integration-constrained model satisfies $\sigma \leq 0.34$, while the two direct-step models give $\sigma = 48.1$ and $\sigma = 60.2$. The two classes are separated by more than two orders of magnitude in the rate at which they amplify perturbations. The leading Lyapunov exponent of the KS attractor at this domain length is of order $10^{-1}$, so the integration-constrained models amplify perturbations at a rate of the same order as the dynamics, while the direct-step models exceed it by a factor of roughly 500. \SItab{1} reports $\sigma$ for every model and \SIsec{1.1} gives the definition.}
% VERIFY BEFORE SUBMISSION: measure the leading Lyapunov exponent of the KS
% system at L=100 from the DNS. If it is near 0.09, the factor of 500 stands
% (48.1/0.09 = 534, 60.2/0.09 = 669). Otherwise adjust or cut the last claim.

For high-dimensional systems the full eigendecomposition of $\mathbf{J}$ can be prohibitive, but the clustering of eigenvalues near one gives a cheap estimate of $|\lambda_{\max}|$. Starting from
\begin{equation}
    \sum_{i=1}^{d}\lambda_i = Tr(\mathbf{J}),
\end{equation}
and the triangle inequality
\begin{equation}
   |Tr(\mathbf{J})| =\left|\sum_{i=1}^{d}\lambda_i\right| \leq \sum_{i=1}^{d}\left|\lambda_i\right|,
   \label{eq:eigen_triangle_inequality}
\end{equation}
the clustering $\lambda_1\approx\lambda_2\approx\cdots\approx\lambda_d\approx\lambda_{\max}$ for integration-constrained models (Fig.~\ref{fig:Fig_1}(c,d), Eq.~(\ref{eq:jacobian_H})) turns Eq.~(\ref{eq:eigen_triangle_inequality}) into an approximate equality, so
\begin{equation}
    |\lambda_{\max}| \approx \frac{1}{d}\,|Tr(\mathbf{J})|,
\end{equation}
with $d$ the dimension of $\mathbf{J}$. The trace can in turn be estimated with computationally tractable random sketching methods~\citep{meyer2021hutch}, avoiding the full spectrum.

\subsection{Deviation from linear stability theory}
\label{sec:deviation}

\begin{figure}
\centering
\includegraphics[width=1.0\linewidth]{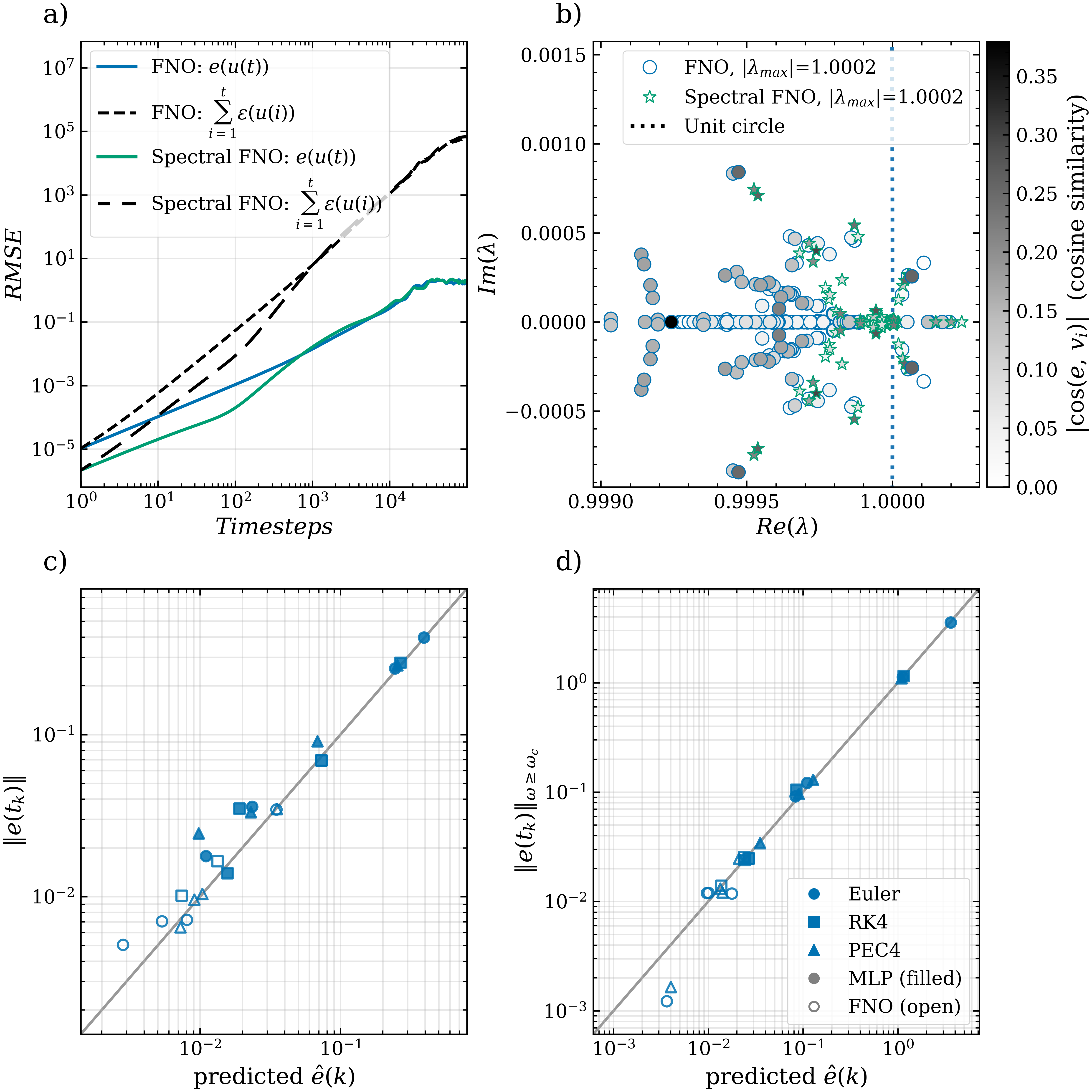}
\caption{\label{fig:Fig_4}
\textbf{Error growth is set by the accumulated generalization error and the error--eigenvector projection, not by $|\lambda_{\max}|$ alone, and is predicted by the linearized error law.} (a, b) Two FNO models constrained with the same PEC4 integrator, trained without (FNO, blue) and with (Spectral FNO, green) the spectral regularizer $\mu(\theta)$ of Eq.~(\ref{eq:spec_loss}). (a) RMSE over the autoregressive rollout on log--log axes. For each model the solid line is the total error $\mathbf{e}(u(t))$ and the black dashed line is the cumulative generalization error $\sum_{i=1}^{t}\boldsymbol{\epsilon}(u(i))$, the leading term of Eq.~(\ref{eq:autoreg_error_time_with_H}), with the dash length distinguishing the two models. The spectrally regularized model has the smaller error over the early rollout. (b) Eigenvalues of the model Jacobian near $\lambda = 1$ for the FNO ($\circ$) and Spectral FNO ($\star$), with the unit circle dotted; to the precision of the legend the two models share $|\lambda_{\max}| = 1.0002$. Each marker is shaded by the cosine similarity $|\cos(\mathbf{e}, \mathbf{v}_i)|$ between its eigenvector $\mathbf{v}_i$ and the error vector $\mathbf{e}(t)$ (grayscale bar). (c, d) The linearized error law of Eq.~(\ref{eq:linearized_error_law}), evaluated with $a = |\lambda_{\max}|$ from the Jacobian at the initial condition, tested after $k = 10$ rollout steps ($t_k = t_0 + k\Delta t$) for every explicit integration-constrained model (marker shape: integration scheme, Euler / RK4 / PEC4; fill: MLP filled, FNO open); each point is one model and the gray line is $y = x$. (c) Total error, with $\|\epsilon\|$ the measured one-step error. (d) The same law restricted to the high-wavenumber band ($\omega \geq \omega_c$, with $\omega_c = 100$), with the injection term replaced by $\beta$, the high-$\omega$ content of the one-step error. See Section~\ref{sec:deviation} for the analysis. \new{The two models in (a, b) are models 11 and 12 of \SItab{1}, and the models in (c, d) are models 1 to 24.}}
\end{figure}

We now revisit the linear scaling law of Section~\ref{sec:linear_scaling} and examine where the eigenvalue-only view breaks down, that is, where a model with a larger $|\lambda_{\max}|$ nonetheless exhibits lower error growth than a model with a smaller one. Figure~\ref{fig:Fig_4}(a,b) compares two FNO models constrained with the same PEC4 integrator, trained with and without the spectral regularizer $\mu(\theta)$. To the precision of the legend the two models share $|\lambda_{\max}| = 1.0002$, with \new{\SItab{1}} separating them only in the fifth decimal place, so the eigenvalue magnitude alone cannot account for the difference in their error growth; yet the spectrally regularized model has the visibly smaller error over the early rollout in Fig.~\ref{fig:Fig_4}(a).

Equation~(\ref{eq:autoreg_error_time_with_H}) identifies two mechanisms behind this deviation. The first is the additive contribution of the generalization error, $\sum_i \boldsymbol{\epsilon}(\mathbf{u}_T(t_i))$. The spectral regularizer, by construction, suppresses the high-wavenumber component of the one-step error and thereby reduces this sum, shown as the dashed lines in Fig.~\ref{fig:Fig_4}(a). Over the early rollout the running sum tracks the total error for each model, confirming that the accumulated generalization error is the leading term of Eq.~(\ref{eq:autoreg_error_time_with_H}); at late times the total error saturates while the running sum continues to accumulate, marking where the small-error approximation underlying the linear theory ceases to hold.

The second mechanism is that the amplification depends on the direction of the error and not only on the eigenvalue magnitudes. The Jacobian acts on the vector $\mathbf{e}(t)$, so the realized growth $\|\mathbf{J}\mathbf{e}\|/\|\mathbf{e}\|$ is a weighted average of the $|\lambda_i|$ over the components of $\mathbf{e}$ in the eigenbasis of $\mathbf{J}$, and it equals $|\lambda_{\max}|$ only when $\mathbf{e}$ is aligned with the leading eigenvector.
When the eigenvalues $\lambda_i$ are tightly clustered around unity, as is characteristic of integration-constrained models, the growth of $\mathbf{e}(t)$ is determined not only by $|\lambda_{\max}|$ but also by the relative orientation (cosine similarity) between $\mathbf{e}(t)$ and the eigenvectors $\mathbf{v}_i$ of $\mathbf{J}$. In Fig.~\ref{fig:Fig_4}(b) each eigenvalue is shaded by $|\cos(\mathbf{e}, \mathbf{v}_i)|$: the largest-magnitude eigenvalues of the spectrally regularized model carry very low cosine similarity, so they contribute little to the realized amplification, and the marginally larger $|\lambda_{\max}|$ does not translate into larger $\|\mathbf{e}(t+\Delta t)\|$. \new{\SIsec{4} shows that these direction-dependent effects act within the $\mathcal{O}(\Delta t)$ residue of the leading-order eigenvalue approximation, so they are finite-$\Delta t$ corrections to the law rather than a failure of it.}

Both mechanisms can be folded into a refined error approximation that remains a priori. Iterating the linearized recursion of Eq.~(\ref{eq:one_step_recursion}) from the initial condition, with the injected one-step error held at its measured norm $\|\boldsymbol{\epsilon}\|$ and the per-step amplification set to the single controlled scale $a = |\lambda_{\max}|$, gives the geometric law
\begin{equation}
    \hat{e}(k) = \|\boldsymbol{\epsilon}\|\,\frac{a^{k}-1}{a-1},
    \label{eq:linearized_error_law}
\end{equation}
where $k$ counts steps from the initial condition, $t_k = t_0 + k\Delta t$, so that $\hat{e}(k)$ is the predicted error after $k$ rollout steps. Note that this differs from the lead time of Fig.~\ref{fig:Fig_2}, which is measured from the first step $t_1$: the reference slope there at lead time $k$ is $p = k+1$, whereas here $\hat{e}(k) \to k\,\|\boldsymbol{\epsilon}\|$ as $a \to 1$, recovering the linear scaling law of Eq.~(\ref{eq:linear_scaling_law}) with $p = k$, and departing from it as $a$ moves away from one. Both inputs, $a$ and $\|\boldsymbol{\epsilon}\|$, are measured at the initial condition, so Eq.~(\ref{eq:linearized_error_law}) predicts the rollout error without running the rollout. Figure~\ref{fig:Fig_4}(c) tests this prediction after $k = 10$ rollout steps for every explicit integration-constrained model: the suite clusters on the line $y = x$, showing that the two-term law of Eq.~(\ref{eq:autoreg_error_time_with_H})---one-step injected error amplified by the Jacobian---predicts error growth across architectures and integration schemes with a single relation. \new{At $k = 10$ and $a = 1.0002$ the geometric factor equals $10.009$, within $0.1\%$ of $k$, so this panel tests the injected-error term with the amplification factor contributing negligibly. The amplification factor becomes discriminating at longer lead times and for the direct-step models, where $a$ departs from unity by five percent.}

The same law also localizes the origin of spectral bias. Restricting the comparison to the high-wavenumber band ($\omega \geq \omega_c$, with $\omega_c = 100$), replacing the injection term by $\beta$, the high-$\omega$ content of the one-step error, and keeping the same amplification $a = |\lambda_{\max}|$, again places the suite on $y = x$ (Fig.~\ref{fig:Fig_4}(d)). The high-$\omega$ error after $k$ rollout steps is therefore set by the high-$\omega$ content of the injected one-step error rather than by any preferential amplification of high wavenumbers, identifying the injected error spectrum $\hat{\epsilon}(\omega)$ as the origin of spectral-bias growth. \new{The Fourier spectra of the predicted state and of its time derivative are shown in \SIfig{3}, and \SIsec{3} examines how $|\lambda_{\max}|$ relates to the implicit diffusivity of the model.}

Deviations from the classical eigenvalue-only prediction therefore emerge from two complementary mechanisms, the direction dependence of error amplification, governed by the eigenvector projections of $\mathbf{J}$, and the cumulative influence of the summed generalization errors, which the spectral regularizer reduces. Once the injected error is measured rather than assumed, the linearized law of Eq.~(\ref{eq:linearized_error_law}) recovers its predictive power across the full explicit integration-constrained suite, and \textcolor{black}{a complete characterization of rollout error growth must account for both the amplification (spectral) and injection (generalization-error) components of Eq.~(\ref{eq:autoreg_error_time})}.

\subsection{Novel stability-promoting loss function}
\label{sec:new_loss_results}

The insight about short-term performance obtained from Section~\ref{sec:deviation} allows us to develop a novel loss function (see Eq.~(\ref{eq:new_loss}) in Section~\ref{sec:new_loss}) that uses the linearized model to minimize $\mathbf{e}(t+\Delta t)$ using the projection of $\mathbf{e}(t)$ on $\mathbf{J}$. 
This constrains both the error as well as the eigenvalues to promote a more stable model. 
We evaluate the new loss on the FNO-based models constrained with each of the three explicit integrators (Euler, RK4, and PEC4), comparing each model trained with the baseline RMSE loss against its counterpart trained with the proposed Jacobian-based loss. 
In Fig.~\ref{fig:Fig_5}(a), we verify that training and inference with the new loss follow the scaling obtained in Section~\ref{sec:linear_scaling}, and that $|\lambda_{\max}|$ diagnoses model performance: for every integrator, the Jacobian-based loss yields a smaller $|\lambda_{\max}|\,\|\mathbf{e}(t)\|$ and a correspondingly smaller $\|\mathbf{e}(t+\Delta t)\|$ than the RMSE baseline. 

Figure~\ref{fig:Fig_5}(b) shows the RMSE over the autoregressive rollout for the same six models. 
To quantify the rate of error growth, the legend reports $\alpha$, the slope of a linear fit of $\log_{10}(\mathrm{RMSE})$ against $\log_{10}$ of the timestep over the first $100$ rollout steps, i.e., the exponent of the power law $\mathrm{RMSE} \propto t^{\alpha}$. 
For every integrator, the model trained with the Jacobian-based loss has a smaller $\alpha$ than its RMSE-trained counterpart (Euler: $0.916$ versus $0.956$; PEC4: $0.927$ versus $0.946$; RK4: $0.965$ versus $0.998$), demonstrating that the new loss slows the growth of error during rollout relative to the basic RMSE loss. 
A reduction in the one-step generalization error alone cannot produce this effect: the leading term of Eq.~(\ref{eq:autoreg_error_time_with_H}) is the running sum of $\boldsymbol{\epsilon}$, so a uniformly smaller $\boldsymbol{\epsilon}$ lowers the log-RMSE curve without changing its slope, provided the one-step error evolves similarly along the trajectory, as in the constant-injection law of Eq.~(\ref{eq:linearized_error_law}). 
The smaller $\alpha$ therefore isolates a reduction in the per-step amplification of the accumulated error, the propagated-error term of Eq.~(\ref{eq:linearize}). 
Consistent with Section~\ref{sec:deviation}, this reduction need not appear in $|\lambda_{\max}|$, which bounds but does not set the realized amplification: \new{\SItab{1}} shows that the Jacobian-based loss leaves the FNO $|\lambda_{\max}|$ essentially unchanged, so for these models the suppression acts through the orientation of the accumulated error relative to the strongly amplified eigendirections of $\mathbf{J}$, whereas for the MLP models the loss also reduces $|\lambda_{\max}|$ directly. 
This reduction in the growth rate is an important result: it shows that the Jacobian-driven amplification of error, which the RMSE objective alone leaves uncontrolled, can be regularized directly during training. 
The reduced growth rate does not by itself guarantee long-term stability however. Several models trained with the Jacobian-based loss (in particular the MLP-based ones) still drift to instability over the longest rollouts because their one-step generalization error $\boldsymbol{\epsilon}$ remains large, consistent with Eq.~(\ref{eq:autoreg_error_time_with_H}), which requires both a near-neutral spectrum and a sufficiently small $\boldsymbol{\epsilon}$ for long-term stability.

\section{Conclusion}

Autoregressive neural emulators are currently trained on a single time step and then used over many time steps: the training objective focuses on the one-step error, while the quantity of interest at inference is the error after hundreds or thousands of steps; and nothing in the training procedure constrains how a one-step error is carried forward. 
We have developed a theoretical framework to address this fundamental disconnect. 
At the core of our approach is a single object, the Jacobian of the learned one-step update map taken with respect to the system state and evaluated after training, Eq.~(\ref{eq:jac_def}). 
Our approach is distinct from the parameter gradients used during training; and, surprisingly, it is rarely examined when these learned models are evaluated.

Linearizing the error dynamics about the true trajectory separates the two contributions to emulation error, Eq.~(\ref{eq:autoreg_error_time}): 
\textit{the one-step generalization error sets the rate at which error is injected, and the error propagator $\mathbf{J}$ sets the rate at which injected error is amplified into the propagated error.} 
Training controls the generalization error and leaves the propagated error free, which is why models with comparable one-step accuracy diverge at different rates during rollout. 
The magnitude of the largest eigenvalue of the Jacobian, $|\lambda_{\max}|$, which governs the growth of a perturbation over one step, is an \textit{a priori} diagnostic of that amplification. 
It requires one automatic-differentiation evaluation at a single state and no rollout. Its definition does not reference the architecture, the integration scheme, or the loss function, and across the suite tested here we observe that its predictive value does not depend on them either.

Emulators fall into two classes according to how the update map is built, and
the classes differ in whether the amplification is controlled at all. A
direct-step model predicts the next state in one shot, so its Jacobian is that
of the network itself, Eq.~(\ref{eq:jac_direct_step}), which carries no
dependence on the time step and places no constraint on the spectrum. An
integration-constrained model instead uses the network $\mathcal{N}$ to
estimate the time derivative and a numerical integrator $\mathbf{H}$ to advance
it, so the Jacobian takes the form $\mathbf{I} + \Delta t\,\nabla\mathbf{H}$,
with $\mathbf{H}$ acting on the output of $\mathcal{N}$,
Eq.~(\ref{eq:jacobian_H}). Its eigenvalues then lie within $\mathcal{O}(\Delta t)$
of unity by construction. The implicit form of the same constraint,
$(\mathbf{I} - \Delta t\,\nabla\mathbf{H})^{-1}$, places an eigenvalue outside
the unit circle only along directions in which the learned tangent operator is
expanding. Converted to a growth rate per unit time, which is the comparison
that is meaningful across models trained at different time steps, the
direct-step models amplify perturbations by orders of magnitude faster than
the integration-constrained models. This separation, rather than the small
difference in the per-step eigenvalue, is what the measured rollouts reflect,
and it accounts for the divergence of direct-step architectures reported
previously~\citep{chattopadhyay2023oceannet,chattopadhyay2023challenges,bi2022pangu,pathak2022fourcastnet, jiang2026hierarchical}.

The near-neutral spectrum of the integration-constrained models also yields a scaling law,
Eq.~(\ref{eq:linear_scaling_law}). The amplification per step lies within
$\mathcal{O}(\Delta t)$ of unity by the structure of
Eq.~(\ref{eq:jacobian_H}), and the injected error is nearly unchanged over the
early rollout, which follows from the continuity of $\tilde{\mathbf{G}}$ over a
small step and which the measured rollouts confirm. The accumulated error
after $p$ steps is then $p$ times the first-step error, so the error grows
linearly in the step count rather than geometrically in the step count, which
is what it does once $|\lambda_{\max}|$ is bounded away from unity.
The law holds across architectures, integration schemes, and loss functions,
within the window in which it is derived.

A near-neutral spectrum is necessary for slow error growth, but it is not sufficient
for long-term stability. Several integration-constrained models drift over the
longest rollouts because their accumulated one-step error is large, consistent
with Eq.~(\ref{eq:autoreg_error_time_with_H}). Two further effects limit the
eigenvalue-only view. The realized amplification is a weighted average of the
eigenvalue magnitudes over the components of the accumulated error in the
eigenbasis of the Jacobian, so it depends on the orientation of that error
relative to the eigenvectors, which the eigenvalues bound but do not set. The
high-wavenumber error after $k$ steps is fixed by the high-wavenumber content
of the injected one-step error rather than by preferential amplification of
high wavenumbers, which locates the origin of spectral-bias growth in the
injected error spectrum rather than in the rollout. Measuring the injected
error and retaining $|\lambda_{\max}|$ as the amplification scale recovers the
prediction, Eq.~(\ref{eq:linearized_error_law}). The same analysis yields a
training objective, Eq.~(\ref{eq:new_loss}), that regularizes the
Jacobian-driven amplification directly and reduces the measured error-growth
exponent for every integrator tested.

The eigenspectrum of the integration-constrained operators differs from that
of a numerical solver for the same equations. The eigenvalues cluster near
unity rather than decaying, which makes $|\lambda_{\max}|$ cheap to estimate
and makes the linearization tractable about a single controlled scale. The
same clustering is why the eigenvector projections matter, since no direction
is strongly preferred by magnitude alone. The direct-step Jacobians do decay,
with most of their spectrum massed near the origin and a few eigenvalues
beyond the unit circle, which is why neither the single-scale linearization
nor the trace estimate applies to them.

Our theory has been developed and tested on the Kuramoto--Sivashinsky equation, a
canonical system for multi-scale chaotic dynamics, and one on which a factorial
sweep over architecture, integrator, and loss is feasible. The Jacobian is
evaluated at the initial condition, and although the reported values are
robust over 100 random initial conditions, the diagnostic describes the local
tangent map rather than its evolution along the attractor. The scaling law holds only for small time steps and over rollouts short enough that the accumulated error remains small. Extending the analysis to the evolution of the Jacobian during rollout, and from the
linear theory to a nonlinear one through random matrix
theory~\citep{rmt4dl_ieee_TR,models_htmu_TR,liao2021hessian,pennington2018emergence} and iterated discrete recurrence relation theory~\citep{HodMah20A_TR}, are natural next
steps, as is testing the diagnostic on higher-dimensional emulators.

Numerical analysis provides established tools for deciding, before a
simulation is run, whether a discretization of a differential equation will be
stable, how fast its error will grow, and how that behavior depends on the
scheme and the time step. Scientific machine learning has no comparable body
of analysis, and stabilization strategies for learned emulators are instead
assembled from heuristics and validated by long
rollouts~\citep{lippe2023pde,stachenfeld2021learned,wikner2022stabilizing}.
The framework developed here is a step toward closing that gap. It takes a
trained network as a discrete dynamical system, identifies the operator that
governs its behavior during inference, and derives from it the quantities that
classical stability analysis provides for numerical schemes. Interpretability
work for scientific neural networks has instead centered on saliency maps,
layer-wise relevance propagation, and related inspection of internal
structure~\citep{molnar2020interpretable}. The diagnostic developed here does
not inspect the network, which is what makes it agnostic to architecture and
loss, and diagnostics of this kind have been productive outside scientific
machine learning~\citep{MM19_HTSR_ICML,MM20_SDM,martin2021predicting}.

\section{Acknowledgements}

AC, PH, and MWM designed the research. CA wrote the computational codes and executed the research. 
Early versions of these codes were developed by AC. AC and CA wrote the paper, and all the authors analyzed the results and edited the manuscript. 
AC and CA acknowledge the support from the National Science Foundation (grant no. 2425667), DARPA APaQus program, the Sloan Foundation, and Schmidt Sciences, LLC. PH is grateful to the National Science Foundation (AGS-531264) and Schmidt Sciences, LLC.
MWM acknowledges DARPA, NSF, the DOE Competitive Portfolios grant, and the DOE SciGPT grant.
Computational resources were provided by NSF ACCESS MTH240019, MTH250006 and NCAR CISL UCSC0008, and UCSC0009. 
The computational codes can be found on GitHub: \url{https://github.com/cainsliewastaken/Spectral_Stability}.

\section{Methods and Systems}

\subsection {Kuramoto--Sivashinsky system}

We conduct our data-driven experiments on a high-dimensional multi-scale chaotic dynamical system, given by the Kuramoto--Sivashinsky (KS) equations. 
The governing equations of the KS system are given by:
\begin{equation}
    \frac{\partial u (x,t)}{\partial t} + u(x,t)\frac{\partial u(x,t)}{\partial x} + \frac{\partial ^4u(x,t)}{\partial x^4} + \frac{\partial^2 u (x,t)}{\partial x^2} =0.
    \label{eq:KS_eq}
\end{equation}
We use a domain length ($L=100$; $x \in \left[-50, 50\right]$) to obtain direct numerical simulations (DNS) of the spatio-temporal evolution of this system using an initial condition given by:
\begin{equation}
    u(x,0) = -\cos\!\left(\frac{2\pi x}{L}\right)\left(1 - \sin\!\left(\frac{2\pi x}{L}\right)\right),
\end{equation}
and periodic boundary conditions. 
We use $1024$ spatial grid points and a $\Delta t_{DNS}=10^{-3}$. 
We train the suite of autoregressive models on $150000$ temporal samples of $u(x,t)$ with $\Delta t= \Delta t_{DNS}$. 
We then conduct emulation with the autoregressive models using a new initial condition outside the training dataset for another consecutive $100000$ time steps. 
Depending on the model, such emulation will either become unstable, unphysical, or, in certain cases, remain stable and physically consistent.

\subsection{Autoregressive models}
\label{sec:AR_models}

In order to build our suite of autoregressive models, we assume that the underlying dynamical system represented by the KS equations is unknown and the discrete dynamics are represented by:
\begin{equation}
    \mathbf{u}(x,t+\Delta t) \approx \underbrace{\mathbf{u}(x,t)+\int_t^{t+\Delta t} \underbrace{\mathbf{F}\left(\mathbf{u}(x,t)\right) d t}_{\mathcal{N}[\circ, \theta]}}_{\mathbf{H}[\circ]}.
    \label{eq:dyn_sys}
\end{equation}
In Eq.~(\ref{eq:dyn_sys}), we describe the general formulation of our autoregressive model, where $\mathcal{N}$ is a neural network, and where $\mathbf{H}$ is generally a numerical integration method implemented inside a differentiable layer. 
Most autoregressive models directly represent $\mathbf{H}[\circ]$ with a deep neural network~\citep{pathak2022fourcastnet,lam2022graphcast,bi2022pangu} (\textit{direct step model}, from now on); while, based on our previous studies~\citep{chattopadhyay2023challenges,chattopadhyay2023oceannet,guan2024lucie}, we introduce a numerical integration-based hard constraint that parameterizes $\mathbf{F}\left(u(x,t)\right)$ with a deep neural network and an \emph{explicitly-implemented differentiable layer} to perform \emph{higher-order integration} in $\mathbf{H}[\circ]$. 

For either the direct step model or the models with integration-based hard constraint, we train the parameters, $\theta$, of the model by minimizing the difference between the predicted and true value of $\mathbf{u}(x,t+\Delta t)$ at each time step in the training dataset. 
The general loss function, without any spectral regularizer (see Section~\ref{sec:loss}), is given by the standard root-mean-square error (RMSE):
\begin{equation}
    L(\theta)= \frac{1}{T}\sum^{t=T}_{t=0}||\mathbf{u}(x,t+\Delta t)- \mathbf{u}(x,t)-\Delta t\mathbf{H}\left[\mathcal{N}\left(\mathbf{u}(x,t), \theta \right)\right]||_2 ,
    \label{eq:L2_loss}
\end{equation}
where $T$ is the total number of time steps (temporal samples) on which the model is trained. 

Next, we describe the few different explicit schemes and the one implicit time integration scheme that has been used, and how it has been implemented in our data-driven autoregressive models.   

\subsection{Explicit integration-based hard constraints}

Here, we describe the few explicit integration-based hard constraints that we have used in this paper. 
We have used a first-order Euler integrator, an RK4 integrator, and a $4^{th}$-order PEC integrator. 
For the explicit schemes, the $\Delta t$ used to train the models and run inference is equal to $\Delta t_{DNS}$, except for the large-$\Delta t$ comparison against the implicit scheme in \new{\SIsec{2.3}}.

For the Euler integration scheme, the predicted value of $\mathbf{u}(x, t+\Delta t)$ from Eq.~(\ref{eq:dyn_sys}) is written as:
\begin{equation}
        \mathbf{u}(x,t+\Delta t) = \mathbf{u}(x,t)+\mathcal{N}[\circ, \theta]\Delta t.
    \label{eq:euler}
\end{equation}
The Euler integrator is similar to learning with the residues~\citep{chen2021generalized}, which has been used in several different studies prior to this (although, in most cases, the $\Delta t$ term is neglected, which results in a non-convergent integration scheme~\citep{krishnapriyan2022learning}). 
The integration-based constraints used in this study are a generalization of learning the residues using rigorous higher-order integrators. 

For the RK4 integrator, we extend Eq.~(\ref{eq:euler}) as:
\begin{align}
k_1 &= \mathcal{N}\left[\mathbf{u}(x,t), \theta\right] \\
k_2 &= \mathcal{N}\left[\mathbf{u}(x,t)+\frac{\Delta t}{2} k_1, \theta\right]  \\
k_3 &= \mathcal{N}\left[\mathbf{u}(x,t)+\frac{\Delta t}{2} k_2, \theta\right]  \\
k_4 &= \mathcal{N}\left[\mathbf{u}(x,t)+\Delta t k_3, \theta\right]  \\
\mathbf{u}(x,t+\Delta t) &=
\mathbf{u}(x,t)+\Delta t\underbrace{\frac{1}{6}\left(k_1+2 k_2+2 k_3+k_4\right)}_{\mathbf{H}[\circ]}.
\label{eq:RK4}
\end{align}
For the $4^{th}$-order PEC method (PEC4), we use the following formulation:
\begin{align}
\Tilde{\mathbf{u}}(x,t+\Delta t) &= \mathbf{u}(x,t) + \Delta t\mathcal{N}[\mathbf{u}(x,t), \theta] \\
\Hat{\mathbf{u}}(x,t+\Delta t) &= \mathbf{u}(x,t) + \frac{\Delta t}{2} \left(\mathcal{N}\left[\mathbf{u}(x,t), \theta\right] + \mathcal{N}\left[ \Tilde{\mathbf u}(x,t+\Delta t), \theta\right]\right)  \\ 
\Bar{\mathbf{u}}(x,t+\Delta t) &= \mathbf{u}(x,t) + \frac{\Delta t}{2} \left(\mathcal{N}\left[\mathbf{u}(x,t), \theta\right] + \mathcal{N}\left[\Hat{\mathbf{u}}(x,t+\Delta t),\theta\right]\right) \\
\mathbf{u}(x,t+\Delta t) &= \mathbf{u}(x,t) + \Delta t \underbrace{\frac{1}{2}\left(\mathcal{N}\left[\mathbf{u}(x,t), \theta\right] + \mathcal{N}\left[\Bar{\mathbf{u}}(x,t+\Delta t),\theta\right]\right)}_{\mathbf{H}\left[\circ\right]}.
    \label{eq:PEC4}
\end{align}
For higher-order integrators such as RK4 or PEC4, we represent the predicted $\mathbf{u}(x,t+\Delta t)= \mathbf{u}(x,t)+\Delta t\mathbf{H}\left[\mathcal{N}\left(\mathbf{u}(x,t),\theta\right)\right]$, where $\mathbf{H}$  encapsulates the scheme as a differentiable operator.

\subsection{Implicit integration-based hard constraint}
\label{sec:implicit_method}

Implicit time-integration schemes have improved stability properties for larger $\Delta t$, as compared with explicit integration schemes~\citep{frank1997stability}. 
In order to implement implicit time-integration as a hard constraint, we use the theory of implicit layers~\citep{kawaguchi2021theory} in deep learning. In this approach, we can use any higher-order scheme as our $\mathbf{H}\left[\circ\right]$ operator. 
The equation involving the time stepper is given by:
\begin{equation}
    \underbrace{\mathbf{u}(x,t+\Delta t)}_{\mathbf{y^{*}}} = \mathbf{u}(x,t) + \Delta t\mathbf{H}\left[ \mathcal{N}\left[\underbrace{\mathbf{u}(x,t+\Delta t)}_{\mathbf{y}^{*}}, \theta \right]\right] ,
    \label{eq:implicit_map}
\end{equation}
where $\mathbf{y}^{*}$ is required to be solved via an implicit layer at each epoch (i.e. each value of $\theta$) in the training process. 
In order to do that, we initialize $\mathbf{y}^{*} = \mathbf{u}\left(x,t\right)$, and we execute a fixed-point iteration to converge to the correct $\mathbf{u}(x,t+\Delta t)$. More details about implicit layers and deep equilibrium models can be found in \citet{kawaguchi2021theory,bai2019deep}. 
For the experiments with the implicit method, the chosen $\Delta t$ ranges from $\Delta t_{DNS}$ up to $100\,\Delta t_{DNS}$ \new{(\SIsec{2.3})}. 
Here, we outline the fixed point iteration in Algorithm~\ref{alg:implicit_scheme} used to train (and run inference with) the implicit numerical integrator. 
\new{Differentiating Eq.~(\ref{eq:implicit_map}) with respect to the state gives $\mathbf{J} = (\mathbf{I} - \Delta t\,\nabla\mathbf{H}[\mathcal{N}(\mathbf{y}^{*};\theta)])^{-1}$, which reduces to Eq.~(\ref{eq:jacobian_H}) to first order in $\Delta t$. \SIsec{2.2} gives the derivation and the resulting expansion conditions.}

\begin{algorithm}[h]
    \caption{Implicit time integration scheme}
    \label{alg:implicit_scheme}
    \begin{algorithmic}[1]
        \State \textbf{Input:} $\epsilon$ \Comment{Convergence threshold, typically $10^{-8}$}

        \State At each epoch, fix $\theta$

        \State Initialize $\mathbf{y}^{*}_{0}=\mathbf{u}(x,t)$, $k = 0$

        \Repeat
            \State $k = k + 1$
            \State $\mathbf{y}^{*}_{k} = \mathbf{u}(x,t) + \Delta t\,\mathbf{H}\left[\mathcal{N}\left(\mathbf{y}^{*}_{k-1},\theta\right)\right]$
        \Until{$||\mathbf{y}^{*}_{k}-\mathbf{y}^{*}_{k-1}||_2 \leq \epsilon$}

    \State \textbf{Output:} $\mathbf{y}^{*}_{k}$ \Comment{$\mathbf{y}^{*}_{k}$ is the predicted value of $\mathbf{u}(x,t+\Delta t)$}
    \end{algorithmic}
\end{algorithm}

\subsection{Neural architectures}
We have used an MLP and an FNO to represent $\mathcal{N}$. 
The MLP has $6$ layers and $2000$ neurons in each layer with a ReLU activation. 
The FNO that has been used has $6$ Fourier blocks and retains $512$ modes in each Fourier block along with a width of $32$. 
We have conducted extensive hyperparameter trials before selecting these sets of hyperparameters for our empirical analysis. 
Each hyperparameter trial optimizes the short-term accuracy of the model.

\subsection{Spectral bias and spectral regularizer}
\label{sec:loss}

Generally, to train autoregressive models, previous studies have used a traditional $L_2$ based loss function, optimizing the difference between the true $\mathbf{u}(x,t+\Delta t)$ and its prediction, $\mathbf{H}\left[\mathcal{N}\left(\mathbf{u}(x,t)\right)\right]$, using Eq.~(\ref{eq:L2_loss}). 
However, for multi-scale dynamical systems, e.g., turbulent flows, a \textit{spectral bias}~\citep{chattopadhyay2023challenges,cao2019towards,wang2024multi} prevents the network from learning the high-wavenumber component of the dynamics. 
This often leads to instability and physical inconsistency in the models. 
Furthermore, spectral bias does not only present itself in the Fourier spectrum of the predicted state, but also in its derivative with respect to time. 
In this paper, the time-derivative is represented by $\mathcal{N}\left(\mathbf{u}(x,t)\right)$ for the explicit methods. 
We propose a spectral regularizer to the $L_2$ loss function presented in Eq.~(\ref{eq:L2_loss}) as:
\begin{equation}
    \mu\left(\theta\right)= \frac{1}{T}\sum^{T}_{t=0}\biggl \lVert\frac{\partial \hat{\mathbf{u}}(x,t)}{\partial t} - \mathbf{\hat{H}}\left[\mathcal{N}\left[\mathbf{u}(x,t)\right]\right]\biggr\rVert_2 ,
    \label{eq:spec_loss}
    \end{equation}
where $\hat{\left[\circ\right]}$ represents the 1D Fourier transform in $x$. 
To approximate the true and predicted time derivatives from the training and predicted data, we use a first-order finite difference. 
The total loss, $L_{total}$, is given by:
\begin{equation}
    L_{total}=L(\theta)+\gamma \mu(\theta),
\end{equation}
where $\gamma$ is the Lagrange multiplier set to 0.10 after significant hyperparameter trials. 
\new{\SIsec{3} reports the effect of this regularizer on the Fourier spectrum of the emulation.}

\subsection{Stability-promoting loss function}
\label{sec:new_loss}

\begin{figure}[ht]
\centering
\includegraphics[width=1.0\linewidth]{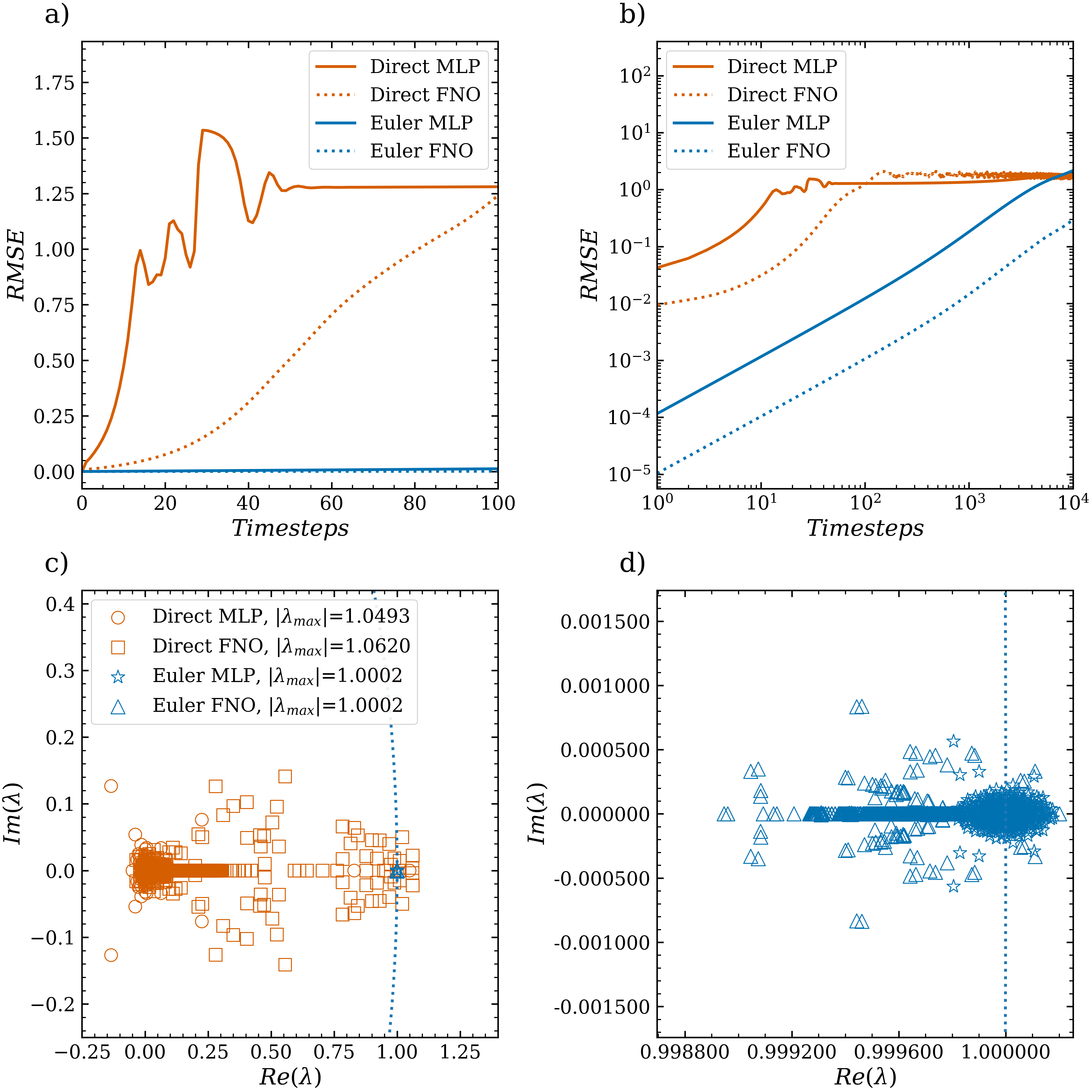}
\caption{\label{fig:Fig_1}
\textbf{Short-term error, rollout error accumulation, and eigenanalysis of the models.} Throughout, direct-step models are shown in vermilion and integration-constrained models in blue, with MLP as solid lines and FNO as dotted lines. 
(a) RMSE over the first $100$ steps. The two direct-step models (Direct MLP, Direct FNO) grow rapidly and saturate near the climatological decorrelation level, where the RMSE is $\mathcal{O}(1)$ in these normalized units. The two integration-constrained Euler models (Euler MLP, Euler FNO), whose eigenvalue structure is representative of all $27$ integration-constrained models examined here, remain stable over this window, with error orders of magnitude smaller and indistinguishable from zero on the linear axis. (b) The same RMSE on log--log axes over the first $10^{4}$ steps. The direct-step models saturate within a few tens of steps (MLP) to a few hundred (FNO), whereas the Euler models grow slowly and in proportion to the step count, \new{remaining orders of magnitude more accurate over most of the window and reaching the decorrelation level only at its right edge}; over the full emulation several MLP-based models eventually drift as their accumulated generalization error grows. (c) Eigenvalues of the model Jacobian $\mathbf{J}$ at the initial condition $t_0$ on the Argand plane for all four models, with $|\lambda_{\max}|$ in the legend. The Direct MLP ($\circ$) and Direct FNO ($\square$) admit eigenvalues with $|\lambda_{\max}| > 1$ ($1.0493$ and $1.0620$) spread well inside and beyond the unit circle, whereas the Euler MLP ($\star$) and Euler FNO ($\triangle$) eigenvalues collapse onto the unit circle (dotted) and cluster tightly at $\lambda = 1$ ($|\lambda_{\max}| = 1.0002$), appearing as a single dense cluster. (d) Zoomed-in view of the integration-constrained eigenvalues from (c), unit circle again dotted. \new{Each spectrum spreads only in the third decimal place, and the two models separate in the fourth}, with $|\lambda_{\max}| = 1.000175$ for the Euler MLP and $1.000201$ for the Euler FNO. This neutral spectrum, consistent across architecture, explains the slow accumulation in proportion to the step count in (a) and (b) through the Jacobian structure of Eq.~(\ref{eq:jacobian_H}). \new{\SIfig{1} shows the equivalent curves and spectra for the remaining explicit models, and \SItab{1} reports $|\lambda_{\max}|$ for every model.}}
\end{figure}

Based on the insight from linear stability analysis as seen in Eq.~(\ref{eq:jacobian_H}), we propose a new stability-promoting loss function for the autoregressive models with integration constraints that minimizes $||\mathbf{e}(t+\Delta t)||_2$ in its linearized form, for each value of $t$ in the training dataset:
\begin{align}
L_{\text{total}} &= \frac{1}{T}\sum_{t=0}^{T}
\biggl\lVert 
\mathbf{J}\Bigl(
\mathbf{u}(t+\Delta t) - \mathbf{u}(t)
- \Delta t\,\mathbf{H}\!\left[\mathcal{N}\!\left[\mathbf{u}(t), \theta \right]\right]
\Bigr)
\biggr\rVert_2
\nonumber \\[4pt]
&\quad +~\mu(\theta)
~+~\frac{1}{T}\sum_{t=0}^{T}
\biggl\lVert 
\mathbf{u}(t+\Delta t) - \mathbf{u}(t)
- \Delta t\,\mathbf{H}\!\left[\mathcal{N}\!\left[\mathbf{u}(t), \theta \right]\right]
\biggr\rVert_2,
\label{eq:new_loss}
\end{align}
where $\mathbf{J}$ is defined in Eq.~(\ref{eq:jacobian_H}).
This novel loss function minimizes the error at the next time step, which is approximated by projecting the predicted state at the next time step on the Jacobian of the model. 
A more aggressive loss function that focuses more on stability could have used the projection of $\mathbf{u}_p(t+\Delta t)$ on the eigenvector of $\mathbf{J}$ corresponding to the largest eigenvalue. 
Such an approximation would have been accurate for a $\mathbf{J}$ with a decaying eigen spectrum; but 
the spectrum of $\mathbf{J}$ for the models with integration constraints is not decaying, instead having eigenvalues that are clustered around $1$ on the unit circle (see Fig.~\ref{fig:Fig_1}(c,d) in Section~\ref{sec:results}). 
As such, projecting $\mathbf{J}$ on the eigenvector corresponding to the largest eigenvalue is often a poor approximation of next time-step error. 
\new{Models trained with the Jacobian term but without the spectral regularizer $\mu(\theta)$ are listed as ``Jacobian'' in \SItab{1}, and those trained with both terms as ``Spectral+Jac.''.}

\begin{figure}
\includegraphics[width=1.0\linewidth]{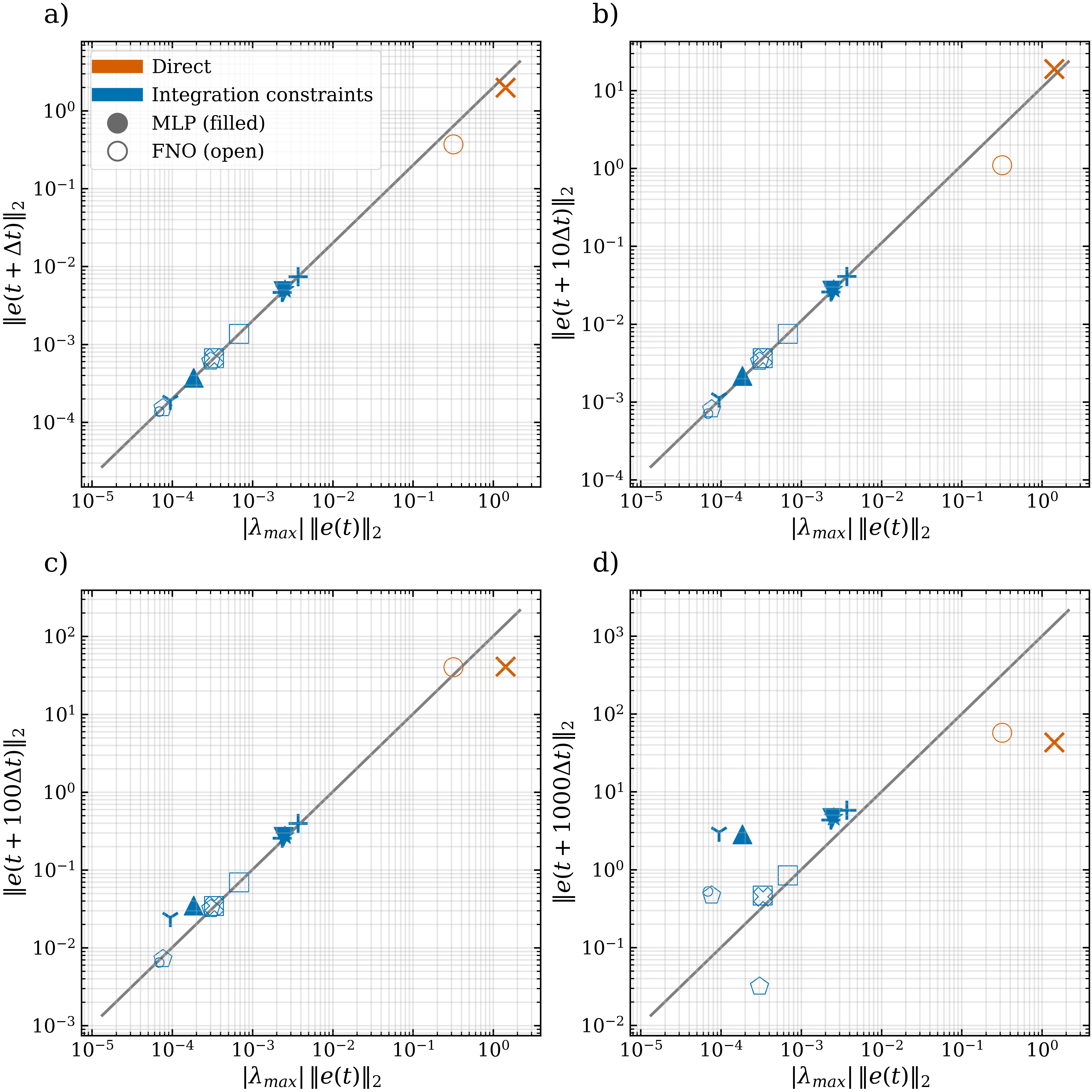}
\centering
\caption{\label{fig:Fig_2}
\textbf{A single linear scaling law governs error growth across the full model suite.} Norm of the accumulated error $\|\mathbf{e}(t + k\Delta t)\|_2$ plotted against $|\lambda_{\max}|\,\|\mathbf{e}(t)\|_2$ for all $29$ models, where $\lambda_{\max}$ is the largest eigenvalue of the model Jacobian $\mathbf{J}$ evaluated at the initial condition and $\|\cdot\|_2$ is the Euclidean ($L_2$) norm of the error vector. Here $t = t_1$, so $\mathbf{e}(t) = \mathbf{e}(t_1) = \boldsymbol{\epsilon}(t_0)$ is the first-step (generalization) error and the lead time $k$ is measured from the first step; the reference slope at lead time $k$ is therefore the step count $p = k+1$ of Eq.~(\ref{eq:linear_scaling_law}). Panels (a)--(d) show the successive lead times $k = 1, 10, 100, 1000$, one per panel, on log--log axes. Marker color denotes model class (direct-step in vermilion, integration-constrained in blue) and fill denotes architecture (MLP filled, FNO open), while marker shape indicates the integration scheme (Euler, RK4, PEC4, implicit Euler). The gray reference line in each panel is the linear scaling law of Eq.~(\ref{eq:linear_scaling_law}) evaluated at that lead time; on log--log axes it is a unit-slope line whose vertical offset grows with the step count, from slope $2$ in (a) to slopes $11$, $101$, and $1001$ in (b)--(d). \new{The panels correspond to $p\,\Delta t = 0.002$, $0.011$, $0.101$, and $1.001$, so they span the validity window of Eq.~(\ref{eq:linear_scaling_law}) from well inside it to its boundary. The integration-constrained models ($|\lambda_{\max}| \approx 1$) lie on the predicted line in (a) and (b), scatter modestly about it in (c), and depart from it in both directions in (d), where $p\,\Delta t \approx 1$ and the derivation no longer applies. The two direct-step models sit at much larger $|\lambda_{\max}|\,\|\mathbf{e}(t)\|_2$ and rise above the reference line at the lead time of (b), reflecting the faster-than-linear growth driven by their $|\lambda_{\max}| > 1$; by (c) and (d) they have reached the climatological decorrelation level and stopped growing, so they fall below a reference line that continues to grow in proportion to $p$.} This \new{faster-than-linear growth} arises from the $\mathcal{O}(\|\mathbf{e}(t)\|_2^2)$ higher-order terms in the Taylor expansion of $\tilde{\mathbf{G}}$, which are neglected in the linear analysis but become non-negligible once the accumulated error grows large. \new{Each model is identified in \SItab{1}. Two of the implicit-Euler models are trained at $\Delta t \gg \Delta t_{\text{DNS}}$ and therefore exceed the $p\,\Delta t \ll 1$ window by a larger margin at the longest lead time (\SIsec{2.4}).}}
\end{figure}

\begin{figure}
\centering
% NOTE: point this \includegraphics at the two-panel stability-promoting-loss image (the linearized-error-law panels now live in Figure_6.png).
\includegraphics[width=1.0\linewidth]{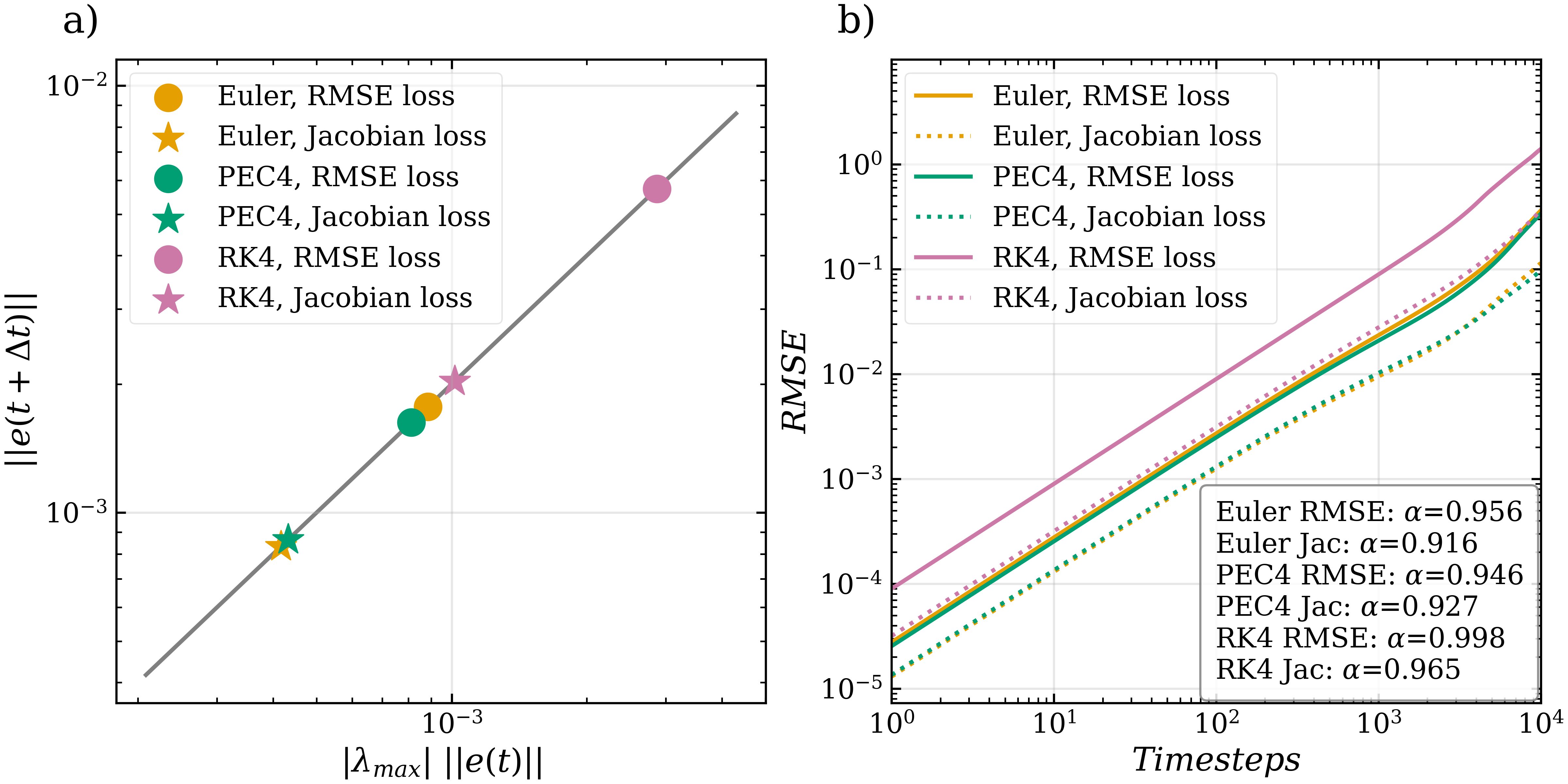}
\caption{\label{fig:Fig_5}
\textbf{The stability-promoting loss lowers error growth.} FNO-based models constrained with the Euler, PEC4, and RK4 integrators, each trained with the baseline RMSE loss and compared against its counterpart trained with the proposed Jacobian-based stability-promoting loss of Eq.~(\ref{eq:new_loss}). (a) The linear scaling law of Fig.~\ref{fig:Fig_2}, $\|\mathbf{e}(t + \Delta t)\|$ versus $|\lambda_{\max}|\,\|\mathbf{e}(t)\|$, with the RMSE baselines shown as circles and the stability-promoting loss as stars. All points lie on \new{the predicted unit-slope line of offset $2$} (gray), confirming that the new loss preserves the linear scaling law, and for every integrator the stability-promoting loss yields a smaller $|\lambda_{\max}|\,\|\mathbf{e}(t)\|$ and correspondingly smaller $\|\mathbf{e}(t + \Delta t)\|$ than the baseline. (b) RMSE over the autoregressive rollout for the same models, with the RMSE baselines shown as solid lines and the stability-promoting loss as dotted lines. The legend reports $\alpha$, the slope of a linear fit of $\log_{10}(\mathrm{RMSE})$ against $\log_{10}$ of the timestep over the first $100$ rollout steps (the exponent of $\mathrm{RMSE} \propto t^{\alpha}$). For every integrator, the stability-promoting loss produces a smaller $\alpha$, and hence lower error growth, than the corresponding RMSE baseline, because it constrains the Jacobian-driven amplification of error, which the RMSE objective alone leaves uncontrolled. \new{The six models shown are models 7, 9, 11, 19, 21, and 23 of \SItab{1}.}}
\end{figure}

%=============================================================================
% ---- SUPPORTING INFORMATION STARTS HERE ------------------------------------
% Everything above is main.tex; everything below is SI_Appendix.tex, verbatim.
% The only thing this file adds is the counter block immediately below.
%
% Section numbering: the section counter restarts, so the SI sections number
% 1 to 4 exactly as in the standalone SI. That is why every pointer must keep
% its "SI," or "of the main text" qualifier; without it, "Section 2.3" is
% ambiguous in this document. All existing pointers already carry one.
%
% Float and equation numbering: SI figures, tables, equations, and algorithms
% keep their S prefix and restart at S1, so they never collide with the
% main-text 1 to 5, 1 to 38, and Algorithm 1.
%
% \mainref is defined here rather than printing a frozen number, because in one
% document the main-text labels resolve. Output is identical either way.
%=============================================================================
\clearpage

\newcommand{\mainref}[2]{\ref{#1}}

\setcounter{section}{0}
\renewcommand{\thefigure}{S\arabic{figure}}
\setcounter{figure}{0}
\renewcommand{\thetable}{S\arabic{table}}
\setcounter{table}{0}
\renewcommand{\theequation}{S\arabic{equation}}
\setcounter{equation}{0}
\renewcommand{\thealgorithm}{S\arabic{algorithm}}
\setcounter{algorithm}{0}

\thispagestyle{empty}

\begin{center}
{\LARGE \textbf{Supporting Information for}}\\[0.8em]
{\Large Eigenanalysis framework for autoregressive neural emulators of multi-scale chaotic dynamics}\\[1.2em]
{\large Conrad Ainslie, Pedram Hassanzadeh, Michael W. Mahoney, and Ashesh Chattopadhyay}\\[1.2em]

E-mail: \texttt{aschatto@ucsc.edu}
\end{center}

\vspace{1.5em}

\noindent\textbf{This PDF file includes:}
\begin{quote}
Supporting text, Sections \ref{si:suite} to \ref{si:normality}\\
Figs. S1 to S3\\
Table S1\\
Algorithm S1\\
SI References
\end{quote}

\vspace{1.5em}
\hrule
\vspace{1.5em}

\section*{Supporting Information Text}
\addcontentsline{toc}{section}{Supporting Information Text}

This document contains four sets of supporting results. Section~\ref{si:suite} reports the full 29-model suite together with its measured Jacobian spectra, and introduces the time-normalized growth rate used to compare models trained at different time steps. Section~\ref{si:implicit} describes the implicit integration-based hard constraint, derives the Jacobian of the implicit update map, and reports the comparison against explicit schemes over a range of $\Delta t$. Section~\ref{si:specbias} presents the Fourier-space analysis of spectral bias. Section~\ref{si:normality} establishes the near-normality of the integration-constrained Jacobian and the resulting validity of the eigenvalue approximation used in Section~\mainref{sec:scaling_derivation}{2.4} of the main text.

%==============================================================================
\section{The full model suite}
\label{si:suite}
%==============================================================================

\begin{figure}[ht]
\centering
\includegraphics[width=1.0\linewidth]{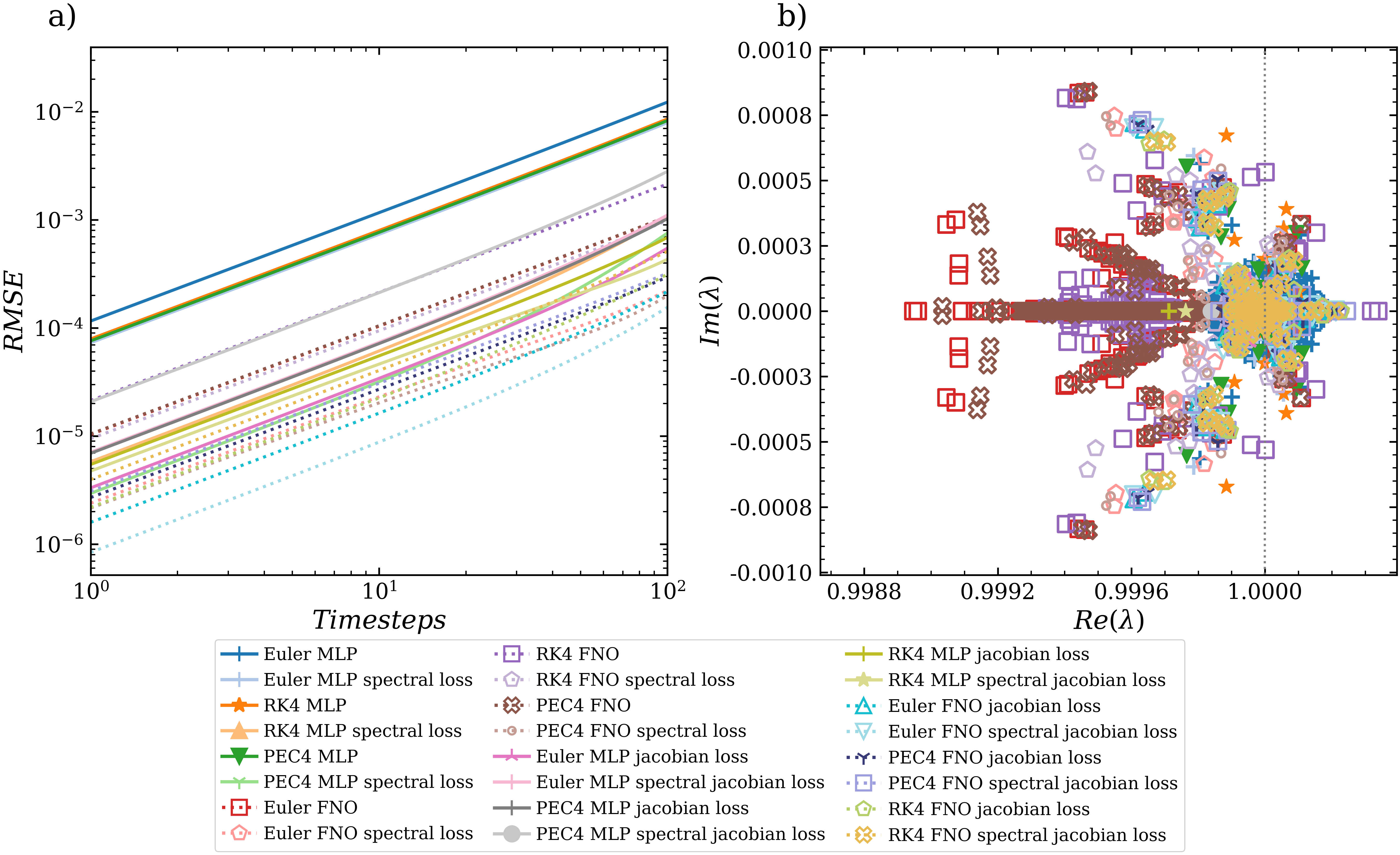}
\caption{\label{fig:S_explicit_suite}
\textbf{Short-term error and eigenspectra of the full explicit
integration-constrained suite.} All 24 explicit integration-constrained models
of Table~\ref{tab:S_model_suite}, spanning the Euler, RK4, and PEC4 schemes,
the MLP and FNO architectures, and the RMSE, spectral, Jacobian, and
spectral+Jacobian losses. Every model is trained and emulated at
$\Delta t = \Delta t_{\text{DNS}} = 10^{-3}$, with MLP models drawn as solid
lines and FNO models as dotted lines, and color and marker distinguishing
individual models as indicated in the legend. (a) RMSE over the first 100
rollout steps on log--log axes. Every model exhibits the slow, near-linear
error accumulation predicted by Eq.~(\mainref{eq:linear_scaling_law}{16}) of the main
text; the curves separate in vertical offset, set by the one-step
generalization error $\boldsymbol{\epsilon}$, rather than in slope, consistent
with the near-neutral spectra shared across the suite. \new{The window shown
satisfies $p\,\Delta t \leq 0.1$, so it lies inside the validity range of that
scaling law.} (b) Eigenvalues of the model Jacobian $\mathbf{J}$, evaluated at
the initial condition $t_0$, on the Argand plane, with the unit circle dotted.
For every model the eigenvalues collapse onto the unit circle and cluster
tightly at $\lambda = 1$, with the spread of each spectrum confined to the
third decimal place and the models separated in the fourth
(Table~\ref{tab:S_model_suite} reports $|\lambda_{\max}|(t_0)$ for each). This
confirms that the qualitative spectrum is insensitive to the architecture, the
explicit integration scheme, and the loss function, as expected from the
Jacobian structure of Eq.~(\mainref{eq:jacobian_H}{10}) of the main text.}
\end{figure}

Table~\ref{tab:S_model_suite} lists all 29 models evaluated in this work. Two are direct-step models and 27 are integration-constrained. Figure~\ref{fig:S_explicit_suite} demonstrates the RMSE growth and the Jacobian eigenspectra of the 24 explicit integration-constrained models in a single view, providing the complete-suite reference behind the representative subsets shown in Figs.~\mainref{fig:Fig_1}{2} and \mainref{fig:Fig_2}{3} of the main text.

\subsection{Time-normalized growth rate}
\label{si:growthrate}

\new{
The largest eigenvalue $|\lambda_{\max}|$ is a per-step quantity, so models trained at different time steps cannot be ranked by $|\lambda_{\max}|$ alone. Interpreting the learned update map as a time-$\Delta t$ flow map, $|\lambda_{\max}| = \exp(\sigma\,\Delta t)$ defines a growth rate per unit time,
\begin{equation}
    \sigma \;=\; \frac{\ln |\lambda_{\max}|}{\Delta t},
    \label{eq:S_growth_rate}
\end{equation}
which is comparable across $\Delta t$ and carries the units of a Lyapunov exponent. Table~\ref{tab:S_model_suite} reports $\sigma$ alongside $|\lambda_{\max}|$.

The normalization separates the two model classes by more than two orders of magnitude. Every integration-constrained model has $\sigma \leq 0.34$, and the majority lie in the range $0.01$ to $0.25$. The two direct-step models have $\sigma = 48.1$ and $\sigma = 60.2$. The direct-step models therefore amplify perturbations at a rate between $140$ and $180$ times larger than the fastest integration-constrained model, and this separation, rather than the difference in the per-step eigenvalue, is what the rollouts of Fig.~\mainref{fig:Fig_1}{2}(a,b) of the main text reflect. The integration-constrained values are of order $10^{-1}$ per unit time, which is the order of the leading Lyapunov exponent of the KS attractor, so these models amplify perturbations at a rate set by the dynamics rather than by the emulator.
% VERIFY BEFORE SUBMISSION: compute the leading Lyapunov exponent of the KS
% system at L=100 from the DNS and compare against the sigma values above.
% Report the measured value here if it supports the statement; soften or cut
% the last sentence if it does not.

The quantity $\sigma$ inherits the precision of $|\lambda_{\max}|$. At $\Delta t = 10^{-3}$ the five-decimal values of Table~\ref{tab:S_model_suite} resolve $\sigma$ to one or two significant figures only, so small differences in $\sigma$ within the integration-constrained class should not be over-interpreted. Equation~(\ref{eq:S_growth_rate}) also uses $|\lambda_{\max}|$ and therefore describes the modulus of the dominant multiplier, not the full complex spectrum.
}

\begin{table}[t]
\centering
\small
\setlength{\tabcolsep}{4pt}
\caption{\label{tab:S_model_suite}\textbf{The full model suite.} All 29 models
evaluated in this work, comprising 27 integration-constrained models and 2
direct-step models. Every explicit model is trained and run at
$\Delta t = \Delta t_{\mathrm{DNS}} = 10^{-3}$; the two large-$\Delta t$
implicit models (26, 27) are the exception.
$|\lambda_{\max}|(t_0)$ is the largest-magnitude eigenvalue of the model
Jacobian at the initial condition.
\new{$\sigma = \ln|\lambda_{\max}|/\Delta t$ is the corresponding growth rate
per unit time of Eq.~(\ref{eq:S_growth_rate}), which is comparable across
$\Delta t$; its precision is limited by the five decimal places reported for
$|\lambda_{\max}|$.}
RMSE is the short-term error at $k=10$ steps. The spectral-bias metric is
$\langle \log_{10}(|\hat{u}_{\mathrm{pred}}|/|\hat{u}_{\mathrm{true}}|)\rangle_{\omega \geq 100}$
at step 100, with positive values indicating excess high-wavenumber energy
relative to truth.}
\begin{tabular}{r l l l c S[table-format=1.5] S[table-format=2.3] c S[table-format=+2.2]}
\toprule
\# & Arch. & Scheme & Loss & $\Delta t$ & {$|\lambda_{\max}|(t_0)$} & {\new{$\sigma$}} & {RMSE ($k=10$)} & {Spec.\ bias} \\
\midrule
1 & MLP & Euler & RMSE & $10^{-3}$ & 1.00018 & 0.180 & \num{1.289e-03} & +12.46 \\
2 & MLP & Euler & Spectral & $10^{-3}$ & 1.00002 & 0.020 & \num{3.850e-04} & +10.46 \\
3 & MLP & RK4 & RMSE & $10^{-3}$ & 1.00021 & 0.210 & \num{1.310e-03} & +12.47 \\
4 & MLP & RK4 & Spectral & $10^{-3}$ & 1.00007 & 0.070 & \num{3.970e-04} & +10.44 \\
5 & MLP & PEC4 & RMSE & $10^{-3}$ & 1.00018 & 0.180 & \num{1.273e-03} & +12.49 \\
6 & MLP & PEC4 & Spectral & $10^{-3}$ & 1.00030 & 0.300 & \num{3.577e-04} & +10.35 \\
7 & FNO & Euler & RMSE & $10^{-3}$ & 1.00020 & 0.200 & \num{1.142e-04} & +9.82 \\
8 & FNO & Euler & Spectral & $10^{-3}$ & 1.00019 & 0.190 & \num{2.526e-05} & +8.79 \\
9 & FNO & RK4 & RMSE & $10^{-3}$ & 1.00034 & 0.340 & \num{2.343e-04} & +9.87 \\
10 & FNO & RK4 & Spectral & $10^{-3}$ & 1.00020 & 0.200 & \num{1.035e-04} & +8.88 \\
11 & FNO & PEC4 & RMSE & $10^{-3}$ & 1.00019 & 0.190 & \num{1.153e-04} & +9.55 \\
12 & FNO & PEC4 & Spectral & $10^{-3}$ & 1.00020 & 0.200 & \num{2.214e-05} & +8.81 \\
13 & MLP & Euler & Jacobian & $10^{-3}$ & 1.00001 & 0.010 & \num{3.850e-04} & +10.86 \\
14 & MLP & Euler & Spectral+Jac. & $10^{-3}$ & 1.00005 & 0.050 & \num{8.121e-05} & +11.01 \\
15 & MLP & PEC4 & Jacobian & $10^{-3}$ & 1.00002 & 0.020 & \num{3.739e-04} & +11.23 \\
16 & MLP & PEC4 & Spectral+Jac. & $10^{-3}$ & 1.00008 & 0.080 & \num{2.335e-04} & +11.03 \\
17 & MLP & RK4 & Jacobian & $10^{-3}$ & 1.00001 & 0.010 & \num{3.935e-04} & +10.96 \\
18 & MLP & RK4 & Spectral+Jac. & $10^{-3}$ & 1.00001 & 0.010 & \num{5.071e-05} & +10.95 \\
19 & FNO & Euler & Jacobian & $10^{-3}$ & 1.00021 & 0.210 & \num{1.794e-05} & +9.83 \\
20 & FNO & Euler & Spectral+Jac. & $10^{-3}$ & 1.00021 & 0.210 & \num{9.718e-06} & +9.80 \\
21 & FNO & PEC4 & Jacobian & $10^{-3}$ & 1.00022 & 0.220 & \num{2.977e-05} & +9.93 \\
22 & FNO & PEC4 & Spectral+Jac. & $10^{-3}$ & 1.00025 & 0.250 & \num{3.381e-05} & +10.27 \\
23 & FNO & RK4 & Jacobian & $10^{-3}$ & 1.00023 & 0.230 & \num{2.482e-05} & +10.00 \\
24 & FNO & RK4 & Spectral+Jac. & $10^{-3}$ & 1.00021 & 0.210 & \num{4.453e-05} & +10.26 \\
25 & FNO & Implicit Euler & RMSE & $10^{-3}$ & 1.00010 & 0.100 & \num{4.131e-04} & +10.21 \\
26 & FNO & Implicit Euler & RMSE & $5 \times 10^{-2}$ & 1.00020 & 0.004 & \num{1.876e-02} & +10.08 \\
27 & FNO & Implicit Euler & RMSE & $10^{-1}$ & 1.01890 & 0.187 & \num{3.840e-02} & +10.48 \\
\addlinespace
\midrule
28 & MLP & Direct & RMSE & $10^{-3}$ & 1.04930 & 48.123 & \num{5.909e-01} & +14.17 \\
29 & FNO & Direct & RMSE & $10^{-3}$ & 1.06200 & 60.154 & \num{3.433e-02} & +11.12 \\
\bottomrule
\end{tabular}
\end{table}

%==============================================================================
\section{Implicit integration-based hard constraints}
\label{si:implicit}
%==============================================================================

\subsection{Implementation}
\label{si:implicit_method}

Implicit time-integration schemes have improved stability properties for larger $\Delta t$, as compared with explicit integration schemes \citep{frank1997stability}. In order to implement implicit time-integration as a hard constraint, we use the theory of implicit layers \citep{kawaguchi2021theory} in deep learning. In this approach, we can use any higher-order scheme as our $\mathbf{H}\left[\circ\right]$ operator. The equation involving the time stepper is given by
\begin{equation}
    \underbrace{\mathbf{u}(x,t+\Delta t)}_{\mathbf{y}^{*}} = \mathbf{u}(x,t) + \Delta t\,\mathbf{H}\left[ \mathcal{N}\left[\underbrace{\mathbf{u}(x,t+\Delta t)}_{\mathbf{y}^{*}}, \theta \right]\right],
    \label{eq:S_implicit_map}
\end{equation}
where $\mathbf{y}^{*}$ is required to be solved via an implicit layer at each epoch, that is, at each value of $\theta$, in the training process. In order to do that, we initialize $\mathbf{y}^{*} = \mathbf{u}\left(x,t\right)$, and we execute a fixed-point iteration to converge to the correct $\mathbf{u}(x,t+\Delta t)$. More details about implicit layers and deep equilibrium models can be found in \citet{kawaguchi2021theory,bai2019deep}. For the experiments with the implicit method, the chosen $\Delta t$ ranges from $\Delta t_{DNS}$ up to $100\,\Delta t_{DNS}$. Algorithm~\ref{alg:S_implicit_scheme} outlines the fixed-point iteration used to train and to run inference with the implicit numerical integrator.

\begin{algorithm}[h]
    \caption{Implicit time integration scheme}
    \label{alg:S_implicit_scheme}
    \begin{algorithmic}[1]
        \State \textbf{Input:} $\epsilon$ \Comment{Convergence threshold, typically $10^{-8}$}
        \State At each epoch, fix $\theta$
        \State Initialize $\mathbf{y}^{*}_{0}=\mathbf{u}(x,t)$, $k = 0$
        \Repeat
            \State $k = k + 1$
            \State $\mathbf{y}^{*}_{k} = \mathbf{u}(x,t) + \Delta t\,\mathbf{H}\left[\mathcal{N}\left(\mathbf{y}^{*}_{k-1},\theta\right)\right]$
        \Until{$||\mathbf{y}^{*}_{k}-\mathbf{y}^{*}_{k-1}||_2 \leq \epsilon$}
    \State \textbf{Output:} $\mathbf{y}^{*}_{k}$ \Comment{$\mathbf{y}^{*}_{k}$ is the predicted value of $\mathbf{u}(x,t+\Delta t)$}
    \end{algorithmic}
\end{algorithm}

\subsection{The Jacobian of the implicit update map}
\label{si:implicit_jacobian}

\new{
The implicit models do not share the Jacobian of the explicitly constrained models, and the difference is what produces their distinct eigenspectrum. Differentiating Eq.~(\ref{eq:S_implicit_map}) with respect to the state $\mathbf{u}_T$, and using the fact that $\mathbf{y}^{*}$ itself depends on $\mathbf{u}_T$, gives
\begin{equation}
    \frac{\partial \mathbf{y}^{*}}{\partial \mathbf{u}_T}
    = \mathbf{I} + \Delta t\,\nabla\mathbf{H}\!\left[\mathcal{N}(\mathbf{y}^{*};\theta)\right]\frac{\partial \mathbf{y}^{*}}{\partial \mathbf{u}_T},
\end{equation}
so that the Jacobian of the implicit update map is
\begin{equation}
    \mathbf{J} = \left(\mathbf{I} - \Delta t\,\nabla\mathbf{H}\!\left[\mathcal{N}(\mathbf{y}^{*};\theta)\right]\right)^{-1}.
    \label{eq:S_implicit_jacobian}
\end{equation}
Expanding Eq.~(\ref{eq:S_implicit_jacobian}) in a Neumann series gives
$\mathbf{J} = \mathbf{I} + \Delta t\,\nabla\mathbf{H}[\mathcal{N}(\mathbf{y}^{*};\theta)] + \mathcal{O}(\Delta t^{2})$,
so the implicit Jacobian agrees with Eq.~(\mainref{eq:jacobian_H}{10}) of the main text to first order in $\Delta t$. The conclusion about explicit schemes carries over to the implicit models, and the implicit models are integration-constrained in that sense. The distinction matters only at $\mathcal{O}(\Delta t^{2})$ and above, which is the regime probed by the large-$\Delta t$ experiments of Section~\ref{si:implicit_model27}.

The inverse structure of Eq.~(\ref{eq:S_implicit_jacobian}) controls the spectrum. Let $\nu_i$ denote the eigenvalues of $\nabla\mathbf{H}[\mathcal{N}(\mathbf{y}^{*};\theta)]$, which carry the units of a rate. The eigenvalues of the implicit Jacobian are
\begin{equation}
    \lambda_i = \frac{1}{1 - \Delta t\,\nu_i},
    \label{eq:S_implicit_eigs}
\end{equation}
whereas the corresponding explicit scheme yields $\lambda_i = 1 + \Delta t\,\nu_i$. Writing $\nu_i = a_i + \mathrm{i}\,b_i$ and evaluating the moduli exactly gives the two conditions for local expansion,
\begin{equation}
    |\lambda_i| > 1
    \;\Longleftrightarrow\;
    a_i > \frac{\Delta t}{2}\,|\nu_i|^{2}
    \quad \text{(implicit)},
    \qquad
    |\lambda_i| > 1
    \;\Longleftrightarrow\;
    a_i > -\frac{\Delta t}{2}\,|\nu_i|^{2}
    \quad \text{(explicit)}.
    \label{eq:S_expansion_conditions}
\end{equation}
The implicit scheme places an eigenvalue outside the unit circle only when the learned tangent operator has an expanding direction, $a_i > 0$. The explicit scheme places an eigenvalue outside the unit circle even for mildly contracting directions with $-\tfrac{\Delta t}{2}|\nu_i|^{2} < a_i < 0$. At fixed $\Delta t$, the implicit spectrum therefore lies closer to and inside the unit disk than the explicit one. This is the origin of the smaller $|\lambda_{\max}|$ and the longer left tail of the implicit eigenspectra in Fig.~\ref{fig:S_implicit}(d), and it is the spectral counterpart of the classical statement that implicit schemes have larger stability regions.
}

\subsection{Short-term and long-term performance}
\label{si:implicit_results}

From classical numerical analysis, we know that implicit numerical time integrators, while more computationally expensive, have larger stability regions than explicit time integration schemes, and in practice are more accurate at large $\Delta t$ for stiff problems. We use our proposed \textit{a priori} diagnostic metric, $|\lambda_{\max}|$, to investigate whether similar properties emerge for neural autoregressive models. 
We consider models that are hard-constrained with implicit and explicit integration schemes and are trained and tested with $\Delta t$ ranging from $\Delta t_{DNS}$ up to $100\Delta t_{DNS}$. To do so, we have performed experiments with $\Delta t = \Delta t_{DNS}=10^{-3}$, $\Delta t = 50\Delta t_{DNS}=5\times 10^{-2}$, and $\Delta t = 100\Delta t_{DNS}=10^{-1}$. Unless otherwise stated, all models in this work are trained and emulated at $\Delta t = \Delta t_{DNS}$. The large-$\Delta t$ models of this section are the exception, and the explicit models trained at $\Delta t > \Delta t_{DNS}$ are used only for this comparison and are not part of the 29-model suite of Table~\ref{tab:S_model_suite}.

\begin{figure}[ht]
\centering
\includegraphics[width=1.0\linewidth]{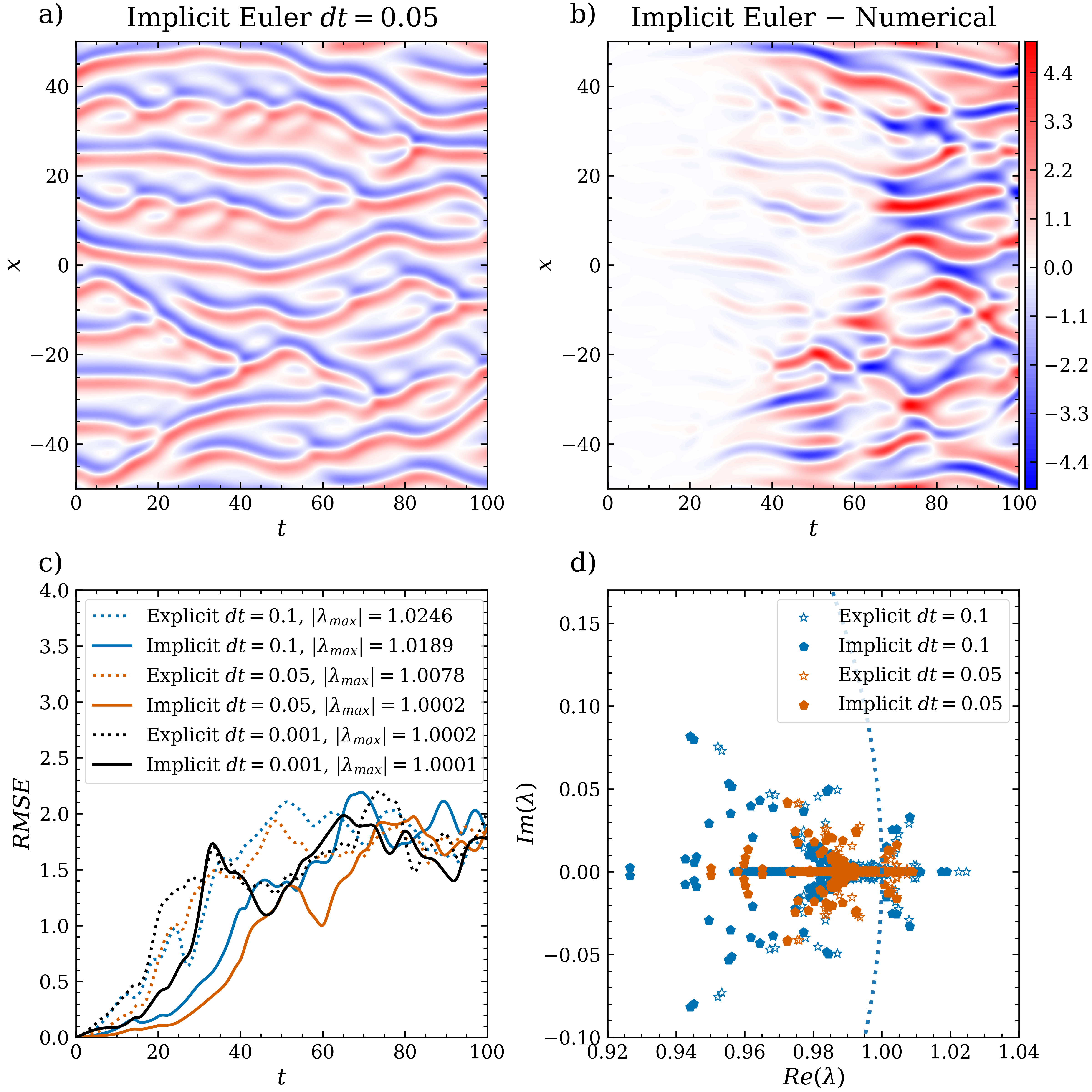}
\caption{\label{fig:S_implicit}
\textbf{Short-term and long-term performance of neural autoregressive models constrained with implicit time integration.} Color encodes the time step, $\Delta t = 0.1$ in blue, $0.05$ in vermilion, $0.001$ in black. Implicit schemes are solid lines and filled pentagons, explicit schemes dotted lines and open stars. (a) Long-term emulation of the KS state $u(x,t)$ by the FNO model constrained with first-order implicit Euler at $\Delta t = 0.05 = 50\,\Delta t_{\text{DNS}}$, stable and physically plausible over $10^{5}$ steps of length $\Delta t_{\text{DNS}}$. (b) Pointwise difference between the emulation in (a) and the reference numerical solution. The error stays small and develops slowly, with structure inherited from the flow rather than diverging, which quantifies the physical consistency. (c) RMSE growth on linear axes for the FNO models with implicit and explicit Euler at the three time steps. At each $\Delta t$ the implicit scheme gives lower error growth, and the linear scale makes the separation visible late in the rollout, where accumulated error dominates. The legend reports $|\lambda_{\max}|$, closer to unity for the implicit scheme in each pair. \new{Within each $\Delta t$ pair the ordering of $|\lambda_{\max}|$ matches the ordering of the RMSE curves, so $|\lambda_{\max}|$ acts as an \textit{a priori} diagnostic of the error-growth rate. Comparison across pairs requires the normalized rate $\sigma$ of Eq.~(\ref{eq:S_growth_rate}), reported in Section~\ref{si:implicit_results}.} (d) Eigenvalues on the Argand plane for both schemes at $\Delta t = 0.1$ and $0.05$, with the unit circle dotted. At each $\Delta t$ the implicit spectrum has a smaller maximum and a more diffusive left tail than the explicit one, consistent with its lower error growth. \new{Equation~(\ref{eq:S_expansion_conditions}) accounts for both. The implicit map places an eigenvalue outside the unit circle only for expanding directions of the learned tangent operator, the explicit map even for mildly contracting ones.}}
\end{figure}

As shown in Fig.~\ref{fig:S_implicit}(a), similar to the explicit integration-based constraint with the Euler scheme, the implicit Euler scheme-based model is also long-term stable and physically consistent. However, the RMSE error curves in Fig.~\ref{fig:S_implicit}(c) clearly demonstrate that for all values of $\Delta t$, the implicit time integration constraint-based models have lower error growth, as compared to explicit ones.
This is expected from traditional numerical analysis. This is further validated by investigating $|\lambda_{\max}|$ of each of the implicit schemes and confirming that they are smaller, that is, closer to unity, than the explicit schemes. Similar to explicit constraints, for implicit constraints as well, $|\lambda_{\max}|$ acts as an \textit{a priori} diagnostic of the error-growth rate of the models. This is of particular importance when developing novel autoregressive architectures for complex systems. The integration constraints allow us to translate our knowledge about the performance and stability of different integration schemes to neural network-based modeling in scientific computing.

\new{The comparison across $\Delta t$ requires the normalization of Eq.~(\ref{eq:S_growth_rate}), since $|\lambda_{\max}|$ is a per-step multiplier and the three experiments use different step sizes. Expressed as growth rates per unit time, the implicit constraint gives $\sigma = 0.100$, $0.004$, and $0.187$ at $\Delta t = 10^{-3}$, $5\times 10^{-2}$, and $10^{-1}$, against $\sigma = 0.200$, $0.155$, and $0.243$ for the corresponding explicit models. The implicit scheme has the lower rate at every $\Delta t$ tested. The raw $|\lambda_{\max}|$ values reported in the legend of Fig.~\ref{fig:S_implicit}(c) order the implicit and explicit models correctly within each $\Delta t$ pair, but they cannot be compared across pairs, and the normalized rate is what makes the three comparisons commensurate.}

\subsection{Large time steps, expanding eigenvalues, and stable emulation}
\label{si:implicit_model27}

\new{
Model 27 of Table~\ref{tab:S_model_suite} is an implicit Euler FNO trained and emulated at $\Delta t = 10^{-1}$. Its Jacobian has $|\lambda_{\max}| = 1.0189$, which places an eigenvalue outside the unit circle, and the linear analysis of Section~\mainref{sec:eig_def}{2.3} of the main text therefore predicts local expansion. The model nonetheless remains stable and physically consistent over the full emulation.

The quantity $|\lambda_{\max}|$ is a multiplier per time step, and models trained at different $\Delta t$ take different numbers of steps to cover the same physical time, so their per-step values cannot be compared with one another. Equation~(\ref{eq:S_growth_rate}) converts $|\lambda_{\max}|$ to a growth rate per unit time, which is comparable. Model 27 gives $\sigma = 0.187$. The two direct-step models give $\sigma = 48.1$ and $\sigma = 60.2$, larger by factors of roughly $260$ and $320$, even though their per-step eigenvalues, $1.049$ and $1.062$, are numerically close to $1.019$. The closeness of the three raw eigenvalues follows from the difference in $\Delta t$ and carries no information about the models.
% Arithmetic: sigma = ln|lambda_max|/dt.
%   model 27: ln(1.0189)/1e-1 = 0.1872
%   direct MLP: ln(1.0493)/1e-3 = 48.123 ; direct FNO: ln(1.0620)/1e-3 = 60.154
%   ratios: 48.123/0.1872 = 257.0 ; 60.154/0.1872 = 321.3

A rate of $\sigma \approx 0.19$ is what an emulator of the KS attractor should give. The attractor has positive Lyapunov exponents, so a tangent operator that reproduces it must have expanding directions, and an emulator whose spectrum lay strictly inside the unit circle would be over-diffusive rather than accurate. Equation~(\ref{eq:S_implicit_eigs}) converts $|\lambda_{\max}| = 1.0189$ at $\Delta t = 10^{-1}$ into an expanding eigenvalue of the learned tangent operator of $0.185$ per unit time, which agrees with $\sigma = 0.187$ to within one percent, the residual difference being of order $\Delta t$ and expected from the two definitions. The expansion described by $|\lambda_{\max}| > 1$ is local and is bounded by nonlinear saturation on the attractor, so it does not imply unbounded divergence, and the emulation confirms this.
% Arithmetic: implicit form lambda = 1/(1 - dt*a) gives a = (1 - 1/1.0189)/1e-1 = 0.1855.
%   (0.1872 - 0.1855)/0.1872 = 0.93 percent.
% VERIFY: leading Lyapunov exponent of KS at L=100, needed to support the
% claim that sigma is of the same order. The main text carries the same check.

Sections~\mainref{sec:eig_def}{2.3} and \mainref{sec:scaling_derivation}{2.4} of the main text assume $\Delta t \ll 1$. At $\Delta t = 10^{-1}$ that condition is marginal, the $\mathcal{O}(\Delta t^{2})$ terms neglected in Section~\ref{si:normality} are no longer small, and the quantitative predictions of the linear theory carry a correspondingly wider tolerance for this model.

Equation~(\mainref{eq:linear_scaling_law}{16}) of the main text holds for $p\,\Delta t \ll 1$, and this restriction governs where the implicit models sit in Fig.~\mainref{fig:Fig_2}{3} of the main text. At the longest lead time, $p = 1001$, models 26 and 27 reach $p\,\Delta t = 50$ and $p\,\Delta t = 100$, so the condition fails by two orders of magnitude. Their departure from the reference line in that panel is what the stated validity window predicts and is not evidence against the scaling law.
% Arithmetic: 1001 * 5e-2 = 50.05 ; 1001 * 1e-1 = 100.1
}

%==============================================================================
\section{Spectral bias and the Fourier spectrum of the emulation}
\label{si:specbias}
%==============================================================================

This section presents the Fourier-space analysis of spectral bias, connecting the eigenvalues of the model Jacobian to the fidelity of the predicted spectrum and its time derivative.

\begin{figure}[ht]
\centering
\includegraphics[width=1.0\linewidth]{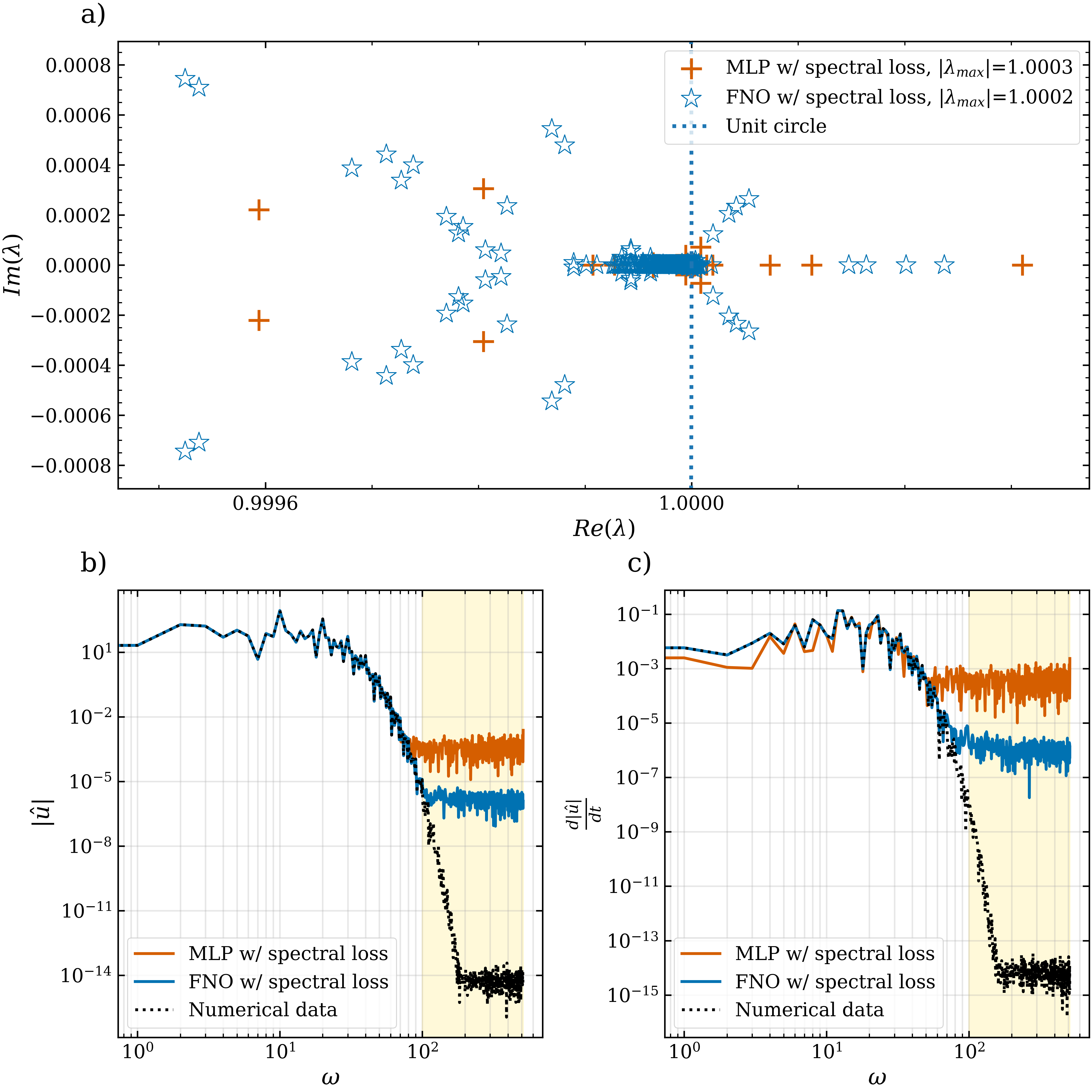}
\caption{\label{fig:S_specbias}
\textbf{Spectral bias and its connection to the eigenvalues of the model Jacobian.} (a) Eigenvalues on the Argand plane for the MLP-based model (vermilion plus signs, $|\lambda_{\max}| = 1.0003$) and the FNO-based model (blue open stars, $|\lambda_{\max}| = 1.0002$), both constrained with the PEC4 integrator and trained with the spectral regularizer of Eq.~(\mainref{eq:spec_loss}{36}) of the main text, with the unit circle dotted. The FNO-based model has the smaller $|\lambda_{\max}|$ and a longer left tail, that is, eigenvalues further inside the unit circle, indicating greater implicit diffusivity. This is reflected in (b) the spatial Fourier spectrum $|\hat{u}(\omega)|$ of the predicted state after $100$ time steps and (c) the spectrum of its time derivative $d|\hat{u}|/dt$. Up to $\omega \approx 100$ both models track the reference numerical data (dotted black). In the shaded high-wavenumber band ($\omega \gtrsim 100$), where the true spectrum continues to decay, both learned models instead saturate at a flat noise floor set by the smallest scale each can represent rather than by the true dynamics. The floor is markedly lower for the FNO (blue) than for the MLP (vermilion) in both the state and its derivative, so the FNO more faithfully captures the high-wavenumber tail. For this matched pair, $|\lambda_{\max}|$ thus tracks the implicit diffusivity and the high-$\omega$ noise floor, with the smaller $|\lambda_{\max}|$ corresponding to greater implicit diffusion and more faithful capture of the high-wavenumber tail of the spectrum.}
\end{figure}

We analyze whether $|\lambda_{\max}|$ can explain different values of spectral bias in neural autoregressive models. As shown previously \citep{chattopadhyay2023challenges,guan2024lucie}, a spectral regularizer in the form of Eq.~(\mainref{eq:spec_loss}{36}) of the main text can reduce spectral bias in neural autoregressive models and improve stability. However, for a decaying spectrum like the one in the KS system, the regularizer does not completely eliminate spectral bias either in the state or in its derivative (Fig.~\ref{fig:S_specbias}(b,c)). This is because the amplitude of the modes of the state in a system with a decaying spectrum decreases as a power law. Figure~\ref{fig:S_specbias}(b) shows that after wavenumber $100$, the slope of the spectrum increases with diminished amplitude of the high-wavenumber modes. Despite the spectral regularizer, spectra with high slopes are more challenging to capture, an effect mirrored in the spectra of state derivatives (Fig.~\ref{fig:S_specbias}(c)) as well. A detailed explanation of the impact of the slope of the system spectrum on the spectral bias of the autoregressive model can be found in \citet{chattopadhyay2023challenges}.

Nonetheless, the extent of spectral bias varies across architectures. For instance, neural operators have been shown to better capture the spatial Fourier spectra of chaotic systems \citep{azizzadenesheli2024neural, oommen2025integrating}. The magnitude of $|\lambda_{\max}|$, representing the distance of the dominant Jacobian eigenvalue from the origin, reflects the implicit diffusivity of a model. Smaller values indicate higher diffusion and often correlate with reduced spectral bias. Figure~\ref{fig:S_specbias}(c) shows that the FNO model equipped with the PEC-based integrator and a spectral regularizer has lower spectral bias, and hence more diffusion, than the MLP-based model. This is captured by the smaller $|\lambda_{\max}|$ of the FNO model as compared to the MLP model, as shown in Fig.~\ref{fig:S_specbias}(a).

The quantity $|\lambda_{\max}|$ therefore complements the one-step error as an \textit{a priori} indicator. It sets the amplification scale, while $\boldsymbol{\epsilon}$ sets the injection. For the matched pair studied here it also tracks the implicit diffusivity and the fidelity of the high-wavenumber spectrum. \new{This correspondence is established for a single matched pair of models and should be read as an empirical association rather than as a general law. Section~\mainref{sec:deviation}{3.3} of the main text establishes the more general statement, namely that the high-wavenumber error after $k$ rollout steps is set by the high-$\omega$ content of the injected one-step error rather than by preferential amplification of high wavenumbers.}

%==============================================================================
\section{Near-normality of the integration-constrained Jacobian and validity of the eigenvalue approximation}
\label{si:normality}
%==============================================================================

The scaling arguments of Section~\mainref{sec:scaling_derivation}{2.4} of the main text replace the action of the Jacobian on the error vector by its largest eigenvalue, $\|\mathbf{J}\,\mathbf{e}\| \approx |\lambda_{\max}|\,\|\mathbf{e}\|$. For a general non-normal matrix this replacement can fail. Eigenvalues bound only the asymptotic growth of perturbations, while transient, orientation-dependent amplification is controlled by the singular values, which for a strongly non-normal matrix can exceed $|\lambda_{\max}|$ substantially. Here we show that the integration-constrained structure, together with the smallness of $\Delta t$, renders the constrained Jacobian near-normal, so that the eigenvalue approximation is accurate to the same order as the theory itself. This argument applies only to the integration-constrained models. No analogous control is available for the direct-step Jacobian.

Write the constrained Jacobian of Eq.~(\mainref{eq:jacobian_H}{10}) of the main text as $\mathbf{J} = \mathbf{I} + \Delta t\,\mathbf{A}$, with $\mathbf{A} = \nabla \mathbf{H}[\mathcal{N}(\mathbf{u}_T;\theta)]$. Its departure from normality is measured by the self-commutator,
\begin{equation}
\mathbf{J}\mathbf{J}^{*} - \mathbf{J}^{*}\mathbf{J} \;=\; \Delta t^{2}\left(\mathbf{A}\mathbf{A}^{*} - \mathbf{A}^{*}\mathbf{A}\right),
\label{eq:S_commutator}
\end{equation}
since the identity commutes with every matrix and the terms linear in $\Delta t$ cancel. The departure from normality is therefore $\mathcal{O}(\Delta t^{2})$, one order smaller than the $\mathcal{O}(\Delta t)$ corrections already retained in the scaling analysis. To first order in $\Delta t$ the constrained Jacobian is normal, regardless of how non-normal the network Jacobian $\mathbf{A}$ itself may be.

The same structure controls the singular values directly. The squared singular values of $\mathbf{J}$ are the eigenvalues of
\begin{equation}
\mathbf{J}^{*}\mathbf{J} \;=\; \mathbf{I} + \Delta t \left(\mathbf{A} + \mathbf{A}^{*}\right) + \Delta t^{2}\,\mathbf{A}^{*}\mathbf{A},
\label{eq:S_gram}
\end{equation}
so that, for bounded $\|\mathbf{A}\|$, every singular value satisfies $\sigma_i(\mathbf{J}) = 1 + \mathcal{O}(\Delta t)$. Consequently, for an error vector of \emph{any} orientation,
\begin{equation}
\frac{\|\mathbf{J}\,\mathbf{e}\|}{\|\mathbf{e}\|} \;\in\; \left[\sigma_{\min}(\mathbf{J}),\, \sigma_{\max}(\mathbf{J})\right] \;=\; 1 + \mathcal{O}(\Delta t),
\qquad\text{while}\qquad
|\lambda_{\max}| = 1 + \mathcal{O}(\Delta t),
\label{eq:S_sv_bracket}
\end{equation}
so that $\|\mathbf{J}\,\mathbf{e}\| = |\lambda_{\max}|\,\|\mathbf{e}\| + \mathcal{O}(\Delta t)\,\|\mathbf{e}\|$. Replacing the vectorial action of $\mathbf{J}$ by the scalar $|\lambda_{\max}|$ is therefore exact at leading order, and the error incurred is of the same $\mathcal{O}(\Delta t)$ size as the corrections already neglected in Section~\mainref{sec:scaling_derivation}{2.4} of the main text. At $\Delta t = \Delta t_{\text{DNS}} = 10^{-3}$ this residue is small, consistent with the fourth-decimal spread of the measured spectra in Fig.~\mainref{fig:Fig_1}{2}(d) of the main text and in Table~\ref{tab:S_model_suite}.

\new{The symbol $\sigma_i$ in Eqs.~(\ref{eq:S_gram}) and (\ref{eq:S_sv_bracket}) denotes a singular value of $\mathbf{J}$ and is distinct from the growth rate $\sigma$ of Eq.~(\ref{eq:S_growth_rate}).}

It is this near-identity structure, rather than the clustering of the eigenvalues per se, that justifies the directional insensitivity invoked in Section~\mainref{sec:scaling_derivation}{2.4} of the main text. Clustered eigenvalues alone would not constrain the transient amplification of a strongly non-normal matrix. The $\mathcal{O}(\Delta t)$ residue left open by the leading-order approximation is the range within which the direction-dependent effects of Section~\mainref{sec:deviation}{3.3} of the main text act. The error-eigenvector projections and the accumulated one-step error modulate the realized amplification within the $\mathcal{O}(\Delta t)$ band around unity. The deviations documented in Section~\mainref{sec:deviation}{3.3} of the main text are thus the expected finite-$\Delta t$ corrections to the leading-order law, rather than a breakdown of the framework. For the direct-step models, by contrast, the Jacobian $\nabla \mathcal{N}(\mathbf{u}_T;\theta)$ carries no $\Delta t$ scaling, its departure from normality is uncontrolled, and eigenvalue-based estimates should be read only as bounds on asymptotic growth.

\new{The same $\Delta t$ scaling delimits where the argument applies. At $\Delta t = 10^{-1}$, used for model 27 of Table~\ref{tab:S_model_suite}, the $\mathcal{O}(\Delta t^{2})$ departure from normality is four orders of magnitude larger than at $\Delta t = \Delta t_{\text{DNS}}$, and the near-normality guarantee is correspondingly weaker. The implicit Jacobian of Eq.~(\ref{eq:S_implicit_jacobian}) admits the same expansion, $\mathbf{J} = \mathbf{I} + \Delta t\,\mathbf{A} + \mathcal{O}(\Delta t^{2})$ with $\mathbf{A} = \nabla\mathbf{H}[\mathcal{N}(\mathbf{y}^{*};\theta)]$, so Eqs.~(\ref{eq:S_commutator}) to (\ref{eq:S_sv_bracket}) hold for the implicit models at leading order as well.}

\clearpage
\bibliographystyle{abbrvnat}
\bibliography{arxiv}

\end{document}